\documentclass[conference]{IEEEtran}

\usepackage{tikz}
\usepackage{amsmath}
\usepackage{amssymb, amsthm}
\usepackage{amsfonts}
\usepackage{comment}
\usepackage{graphicx}
\usepackage{adjustbox}
\usepackage{booktabs}
\usepackage{bbm}
\usepackage{subcaption}

\newtheorem{theorem}{Theorem}
\newtheorem{assumption}{Assumption}

\newtheorem{proposition}{Proposition}

\usepackage{mathtools} 

\ifCLASSINFOpdf
\else
\fi
\begin{document}
%
\title{Temperature Scaling Attack\\Disrupting Model Confidence in Federated Learning}

\author{\IEEEauthorblockN{Kichang Lee and Jaeho Jin}
	\IEEEauthorblockA{Yonsei University\\
		\{kichang.lee, jaeho.jin.eis\}@yonsei.ac.kr}
    \and
	\IEEEauthorblockN{JaeYeon Park}
	\IEEEauthorblockA{Dankook University\\
		jaeyeon.park@dankook.ac.kr}
    \and
	\IEEEauthorblockN{Songkuk Kim and JeongGil Ko}
	\IEEEauthorblockA{Yonsei University\\
		\{songkuk, jeonggil.ko\}@yonsei.ac.kr}}
	

%


\IEEEoverridecommandlockouts
\makeatletter\def\@IEEEpubidpullup{6.5\baselineskip}\makeatother
\IEEEpubid{\parbox{\columnwidth}{
		Network and Distributed System Security (NDSS) Symposium 2027\\
		22--26 March 2027, Seoul, Republic of Korea\\
		ISBN 978-1-970672-09-1\\  
		https://dx.doi.org/10.14722/ndss.2027.[23$|$24]xxxx\\
		www.ndss-symposium.org
}
\hspace{\columnsep}\makebox[\columnwidth]{}}

\maketitle

\begin{abstract}
Predictive confidence serves as a foundational control signal in mission-critical systems, directly governing risk-aware logic such as escalation, abstention, and conservative fallback. While prior federated learning attacks predominantly target accuracy or implant backdoors, we identify \emph{confidence calibration} as a distinct attack objective. We present the \emph{Temperature Scaling Attack (TSA)}, a training-time attack that degrades calibration while preserving accuracy. By injecting temperature scaling with learning rate--temperature coupling during local training, TSA shifts model confidence while keeping predictive accuracy and common optimization signals close to benign training. We provide a convergence analysis under non-IID settings, showing that the coupling controls the primary update scale while leaving a bounded temperature-induced residual, yielding the standard non-convex FL convergence structure with an additional residual term. Across three benchmarks, TSA substantially shifts calibration (e.g., 145\% error increase on CIFAR-100) with ${<}2\%$ accuracy change, and remains effective under robust aggregation and post-hoc calibration defenses. Case studies further show up to a 7.2$\times$ increase in missed verifications in healthcare and severe confidence-gating failures in autonomous driving, even when accuracy is unchanged. Overall, our results establish calibration integrity as a critical attack surface in federated learning.
\end{abstract}

\IEEEpeerreviewmaketitle
\section{Introduction}
\label{sec:intro}

Predictive confidence is a cornerstone of operational reliability in modern safety-critical systems~\cite{park2023self}. In domains such as autonomous driving~\cite{antonante2023task,kendall2017uncertainties}, remote clinical monitoring~\cite{park2020heartquake}, and industrial robotics~\cite{lee2026iros}, probability outputs are not merely explanatory; they are consumed as foundational control signals for risk-aware policies~\cite{kompa2021second}. For example, a vehicle may engage conservative fallback behaviors when confidence is low, and a healthcare pipeline may escalate uncertain cases for human review~\cite{geifman2017selective}. These mechanisms implicitly assume \emph{calibration}: a confidence score should correspond to the empirical likelihood of correctness~\cite{niculescu2005predicting}. When this alignment is systematically distorted, over-confidence can suppress vital safeguards, while under-confidence can trigger excessive verification, degrading both safety and operational efficiency.

In many such systems, federated learning (FL) is increasingly adopted to reduce the need for centralizing sensitive user data while enabling personalization at scale. However, decentralization fundamentally enlarges the adversarial surface. Existing FL security research has largely remained accuracy-centric, focusing on poisoning attacks that disrupt convergence or backdoors that induce targeted misclassifications~\cite{fang2020local,bagdasaryan2020backdoor,xie2019dba}. While effective at causing errors, these attacks can often be detectable in practice~\cite{lee2024tazza,blanchard2017machine,lee2024detrigger}; noticeable drops in validation accuracy or abnormal training dynamics serve as immediate red flags, triggering server-side auditing and mitigation.

For a sophisticated adversary, a more insidious threat is to preserve model utility while subverting its operational reliability~\cite{zhu2022rethinking}.
By targeting to attack the calibration rather than accuracy, the attacker can keep the model deployable and continue to deliver seemingly correct predictions to end users, while systematically distorting the probability scale that downstream policies rely on~\cite{obadinma2024calibration}.
This creates a silent failure mode, given that deployed systems typically monitor only accuracy or loss, as calibration assessment requires representative validation data that is rarely available post-deployment. Consequently, the system may pass standard accuracy-based validation while safety gating and triage logic driven by model confidence become systematically unreliable.

Such a risk is amplified in federated settings with inherently non-IID client data~\cite{mcmahan2017communication}.
Post-hoc calibration is also challenging to apply: each client's validation set is small, biased, and often mismatched to the global deployment distribution~\cite{chen2024watch,peng2024fedcal}.
As a result, local corrections are weakened and ``normal'' confidence behavior becomes ill-defined across clients and rounds, hindering reliable auditing of calibration manipulation.

The consequences of such silent calibration failures can be severe. In autonomous driving, an over-confident model may suppress critical safety fallbacks even when making incorrect predictions, directly increasing accident risk. In healthcare triage, systematic under-confidence can overwhelm clinicians with false alarms, degrading both efficiency and patient safety. Unlike accuracy-based attacks that trigger immediate red flags, calibration manipulation remains deployable while corrupting the signals that safety mechanisms depend on.

However, manipulating confidence during federated training presents a fundamental challenge. Changing the training-time probability mapping also changes the gradients that drive local optimization. Aggressive confidence manipulation can therefore perturb the optimization trajectory and degrade predictive accuracy, defeating the goal of altering calibration without substantially changing model utility. The key challenge is to induce a persistent and directional shift in model confidence while limiting the accompanying change in the underlying optimization dynamics. This challenge is further complicated by non-IID client data, where local updates already vary substantially across clients and communication rounds.

To address this challenge, we introduce the \emph{Temperature Scaling Attack (TSA)}, a training-time mechanism that manipulates model confidence through temperature-scaled local optimization. TSA couples the training temperature with the local learning rate so that the explicit temperature-dependent change in gradient scale is controlled, while the change in the predictive distribution remains able to steer confidence. Our non-convex convergence analysis formalizes this separation and characterizes the remaining temperature-induced deviation from benign optimization.

Across three benchmarks, TSA induces substantial and directional confidence shifts while keeping predictive accuracy close to the benign baseline. We further characterize how these changes appear under common training and update-level monitoring signals. Compared with conventional poisoning baselines, TSA produces substantially smaller changes in update direction and learned representations, showing that calibration manipulation can remain weakly expressed in signals traditionally used to identify model poisoning.

We further evaluate TSA against existing defenses and show that the attack persists under byzantine-robust aggregation rules and post-hoc calibration procedures. Finally, we validate real-world relevance through three case studies. In mobile healthcare triage, confidence manipulation causes up to 7.2$\times$ increase in missed critical cases (over-confidence) or unnecessary clinician reviews (under-confidence). In autonomous driving, miscalibration leads to detection failures that either miss objects entirely or flood the pipeline with false alarms. For language generation, degraded probability estimates corrupt decoding logic, producing repetitive and factually inconsistent outputs. Across all cases, standard accuracy metrics remain largely unchanged, confirming that calibration integrity is a distinct and consequential attack surface. Our core contributions are as follows:

\noindent{}\textbf{Calibration integrity as a new attack surface:} We identify predictive confidence calibration as a standalone attack objective in FL and demonstrate that it can be manipulated without triggering accuracy-centric alarms.

\noindent{}\textbf{Attack design and theoretical foundation:} We propose the $\eta$--$\tau$ coupling mechanism and develop a non-convex convergence analysis under non-IID conditions, explaining how the coupling controls the primary update scale while bounding the remaining temperature-induced deviation that drives confidence manipulation.


\noindent{}\textbf{Behavior under existing monitoring:}
We characterize how TSA appears under common accuracy-, update-, and representation-based monitoring signals. Compared with conventional poisoning baselines, TSA causes substantially smaller changes in update direction and learned representations while still shifting model calibration. These results help explain why defenses designed around accuracy degradation or anomalous update geometry may have limited sensitivity to calibration-oriented attacks.

\noindent{}\textbf{Extensive evaluation and case studies:} We validate TSA across various benchmarks and FL settings, demonstrate robustness against existing defenses, and demonstrate real-world implications through case studies in healthcare, autonomous driving, and language generation pipelines.

\section{Background and Related Work}
\label{sec:background}
\subsection{Security Threats in Federated Learning}
\label{subsec:fl-threats}
The primary security surface in federated learning (FL) arises from malicious clients that can submit adversarial updates to the server during the collaborative training process. We note that we focus on client-side threats and do not consider server-side data inference attacks. Prior work has studied several families of attacks across diverse FL settings.

(i) \textit{Untargeted model poisoning.}
Adversaries perturb local updates to degrade global accuracy or slow convergence, using techniques such as sign flipping, scaled gradients, or noise injection~\cite{bhagoji2019analyzing,lee2024tazza}. These attacks aim to disrupt the learning process and render the global model unusable.
(ii) \textit{Backdoor and targeted poisoning.}
Attackers train on poisoned data so that inputs containing specific triggers are mapped to attacker-chosen labels while maintaining accuracy on clean data~\cite{bagdasaryan2020backdoor,xie2019dba,lee2024detrigger}. Model replacement attacks amplify this by scaling malicious updates to dominate the aggregation, effectively overwriting the global model with a backdoored version.

\textit{Byzantine-robust defenses and evasion.}
To counter malicious updates, robust aggregation mechanisms have been proposed, including Trimmed Mean~\cite{yin2018byzantine}, Krum~\cite{blanchard2017machine}, FLAME~\cite{nguyen2022flame}, and FLTrust~\cite{cao2020fltrust}. In response, sophisticated adversaries have developed collusion strategies to evade these defenses by carefully shaping their updates to appear benign during the aggregation process~\cite{fang2020local}.

Collectively, these lines of work optimize for correctness degradation or backdoors. While effective at causing visible failures, such attacks are often detected through accuracy-based monitoring or abnormal training dynamics. Moreover, the integrity of model \emph{confidence} has received little attention despite its critical role in operational decisions, such as selective prediction, escalation logic, and human-in-the-loop review. Our work positions confidence calibration as an orthogonal threat dimension that can evade accuracy-centric monitors and standard robust aggregation while causing harm in confidence-dependent systems.

\subsection{Model Confidence and Calibration}

Most modern neural networks produce confidence estimates by converting real-valued scores into probabilities via the softmax function.
Given an input sample $x$, a classifier outputs a logit vector $z(x)\in\mathbb{R}^{C}$ over $C$ classes, and the predicted class distribution is
\begin{equation}
p_\theta(y=c\mid x)
=
\frac{\exp(z_c(x))}{\sum_{k=1}^{C}\exp(z_k(x))},
\quad c\in\{1,\ldots,C\}.
\label{eq:softmax}
\end{equation}
The predicted label is $\hat{y}(x)=\arg\max_{c} p_\theta(y=c\mid x)$, and the associated confidence is $\hat{p}(x)=\max_{c} p_\theta(y=c\mid x).$
These probabilities are often treated as measures of model certainty and frequently drive downstream system behavior beyond evaluation purposes.

\textit{Confidence in operational systems.}
In safety-critical deployments, confidence is directly consumed by operational logic.
For instance, in remote patient monitoring, on-device models escalate only low-confidence predictions to clinicians, avoiding continuous data upload.
In autonomous driving, perception confidence gates conservative fallback behaviors.
System correctness and resource allocation thus depend critically on whether predicted probabilities faithfully reflect empirical correctness, a property known as \emph{calibration}.

\textit{Calibration metrics.}
The reliability of predicted probabilities is quantified using calibration metrics.
Let $\{(x_i,y_i)\}_{i=1}^{n}$ be an evaluation set, and define
$\hat{y}_i=\arg\max_{c} p_\theta(y=c\mid x_i)$,
$\hat{p}_i=\max_{c} p_\theta(y=c\mid x_i)$, and
$a_i=\mathbbm{1}[\hat{y}_i=y_i]$ (correctness indicator).
Expected Calibration Error (ECE) measures the discrepancy between confidence and accuracy by partitioning samples into $M$ bins $\{B_m\}_{m=1}^{M}$ according to $\hat{p}_i$:
\begin{equation}
\mathrm{ECE}
=
\sum_{m=1}^{M}\frac{|B_m|}{n}\,
\left|\mathrm{acc}(B_m)-\mathrm{conf}(B_m)\right|,
\label{eq:ece}
\end{equation}
where $\mathrm{acc}(B_m)=\frac{1}{|B_m|}\sum_{i\in B_m} a_i$ and $\mathrm{conf}(B_m)=\frac{1}{|B_m|}\sum_{i\in B_m} \hat{p}_i$ denote per-bin empirical accuracy and average confidence, respectively~\cite{guo2017calibration}.
While ECE measures the magnitude of miscalibration, it discards directionality.
This distinction matters: overconfidence suppresses escalation despite incorrect predictions, while underconfidence triggers unnecessary verification.
Signed-ECE (sECE) retains this direction~\cite{kim2024curved}:
\begin{equation}
\mathrm{sECE}
=
\sum_{m=1}^{M}\frac{|B_m|}{n}\,
\left(\mathrm{acc}(B_m)-\mathrm{conf}(B_m)\right),
\label{eq:sece}
\end{equation}
where $\mathrm{sECE}<0$ indicates overconfidence and $\mathrm{sECE}>0$ indicates underconfidence.
To build intuition, consider weather forecasting: when a service reports ``30\% chance of rain,'' calibration requires that it actually rains on roughly 30\% of such forecasts.
ECE quantifies the average deviation from this ideal, while sECE reveals whether the model systematically over- or under-predicts its correctness likelihood.

Complementary proper scoring rules evaluate probabilistic predictions without binning.
Negative Log-Likelihood (NLL) is defined as
\begin{equation}
\mathrm{NLL}
=
-\frac{1}{n}\sum_{i=1}^{n}\log p_\theta(y=y_i\mid x_i),
\label{eq:nll}
\end{equation}
which heavily penalizes confident mistakes.
The Brier score measures the squared error between the predicted distribution and the one-hot label $e(y_i)\in\{0,1\}^{C}$:
\begin{equation}
\mathrm{Brier}
=
\frac{1}{n}\sum_{i=1}^{n}\sum_{c=1}^{C}
\left(p_\theta(y=c\mid x_i)-e(y_i)_c\right)^2.
\label{eq:brier}
\end{equation}
In this work, we utilize ECE and sECE as primary metrics given their interpretable, deployment-oriented view of probability quality and their ability to distinguish between overconfidence and underconfidence, a distinction critical for understanding how calibration manipulation affects safety-critical decision logic.

\subsection{Attacking Confidence Calibration}
\label{subsec:calibattacks}
Prior work on attacking model calibration has largely focused on \emph{inference-time} adversarial examples that craft input perturbations to induce miscalibration or unreliable uncertainty estimates~\cite{obadinma2024calibration, galil2021disrupting, ledda2023adversarial}.
These attacks typically modify inputs to alter the posterior distribution (e.g., confidence or entropy) without changing the predicted label, often requiring white-box access to model internals (architecture, parameters, logits) or repeated black-box queries~\cite{goodfellow2014explaining,madry2017towards,carlini2017towards}.
While effective in controlled settings, such approaches face two fundamental challenges in real-world deployments.

First, they assume the attacker can compute or estimate sensitive model signals, which may be unavailable due to API restrictions, output rounding, rate limits, or privacy constraints.
Second, even when perturbations can be generated, their effectiveness degrades substantially in physical deployments due to sensing noise, compression artifacts, domain shifts, and non-differentiable pre-processing~\cite{kurakin2018adversarial,eykholt2018robust,lee2025mind,guo2024invisible,lee2023exploiting}.
These challenges are amplified in federated learning, where adversaries cannot control other clients' inputs or consistently access the global model's gradients at inference across devices.
Consequently, inference-time calibration attacks do not naturally extend to FL threat models that operate through local training.

In contrast, \emph{training-time} manipulation of model confidence has received limited attention.
Most FL attacks focus on degrading accuracy, targeted misclassification, or implanting backdoors~\cite{bagdasaryan2020backdoor,fang2020local,lyu2022privacy}, objectives orthogonal to calibration integrity that can often be detected via accuracy monitoring or misprediction patterns.
Our work addresses this gap by treating calibration as a first-class attack objective in FL.
We propose a training-time mechanism that systematically biases the model's confidence through temperature-controlled local optimization, without requiring inference-time perturbations or access to other clients' data.

\subsection{Temperature Scaling}
\label{subsec:ts}
For classification, a model produces a logit vector $z(x)\in\mathbb{R}^{C}$, which is converted into a categorical probability distribution through the softmax function (Eq.~\ref{eq:softmax}). Temperature scaling modifies the sharpness of this distribution by rescaling the logits with a positive scalar
$\tau>0$:
\begin{equation}
p_{\theta}^{(\tau)}(y=c\mid x)
=
\frac{\exp(z_c(x)/\tau)}
{\sum_{j=1}^{C}\exp(z_j(x)/\tau)}.
\end{equation}
Larger temperatures ($\tau>1$) produce smoother, less peaked distributions, whereas smaller temperatures ($0<\tau<1$) produce sharper distributions. Throughout this work, $\tau<1$ refers to $0<\tau<1$.

Temperature has been used in several contexts, including knowledge distillation~\cite{hinton2015distilling}, post-hoc calibration~\cite{guo2017calibration}, and probabilistic sampling~\cite{jang2016categorical,zhu2024hot,zhang2024edt}. 
Beyond changing probabilities directly, prior work has shown that temperature can also affect optimization when applied during training. 
Agarwala et al.~\cite{agarwala2020temperature}, for example, analyze the interaction between softmax temperature, optimization, and generalization under softmax-cross-entropy training.

Training-time temperature scaling has also been studied in federated learning as a means of improving local optimization and generalization~\cite{lee2026improving}.
Our setting uses the same training-time control in an adversarial direction.
In particular, we couple the local learning rate and temperature through $\beta=\eta/\tau$. As shown in Section~\ref{sec:theory}, this coupling removes the explicit $1/\tau$ factor in the softmax-cross-entropy gradient, while a temperature-dependent residual remains because the predictive distribution itself changes.
This separation allows the primary update scale to remain comparable to benign training while temperature still alters model confidence, which forms the basis of TSA.

\section{Threat Model}
\label{sec:threat}
\noindent\textbf{Goals and Capability.}
We define the adversary's goal as manipulating the model confidence calibration toward either under- or over-confidence states, while keeping predictive accuracy ostensibly unchanged. 
For capability, we assume the adversary has full control over the local training process on its own compromised devices.
Concretely, the attacker can choose the training temperature and adjust the local learning rate accordingly during standard local optimization.
We note that these capabilities are standard in the FL security literature~\cite{wen2023survey}.

\noindent\textbf{Assumptions.}
We assume that attackers can control local training but cannot interfere with server-side aggregation or benign clients, and that the fraction of malicious participants is less than $50\%$~\cite{fang2020local}. The adversary can not manipulate other clients' training process or data, and has no control over any computation performed at the server.
\section{Temperature Scaling Attack}
\label{sec:design}
In this section, we present the design of the proposed \emph{temperature scaling attack}. We begin by developing an analytic rationale for temperature scaling from a decision-boundary perspective, clarifying how it can substantially reshape predictive confidence while largely preserving the underlying decision regions. Building on this interpretation, we describe the attack mechanism, which injects carefully regulated temperature scaling into local training to steer confidence and calibration with minimal impact on accuracy. 
Finally, we provide a convergence-based analysis that explains why the resulting model updates can remain difficult to distinguish from benign updates while still inducing systematic miscalibration.

\subsection{Rationale}
\label{sec:rationale}
\begin{figure}
\centering
\includegraphics[width=\linewidth]{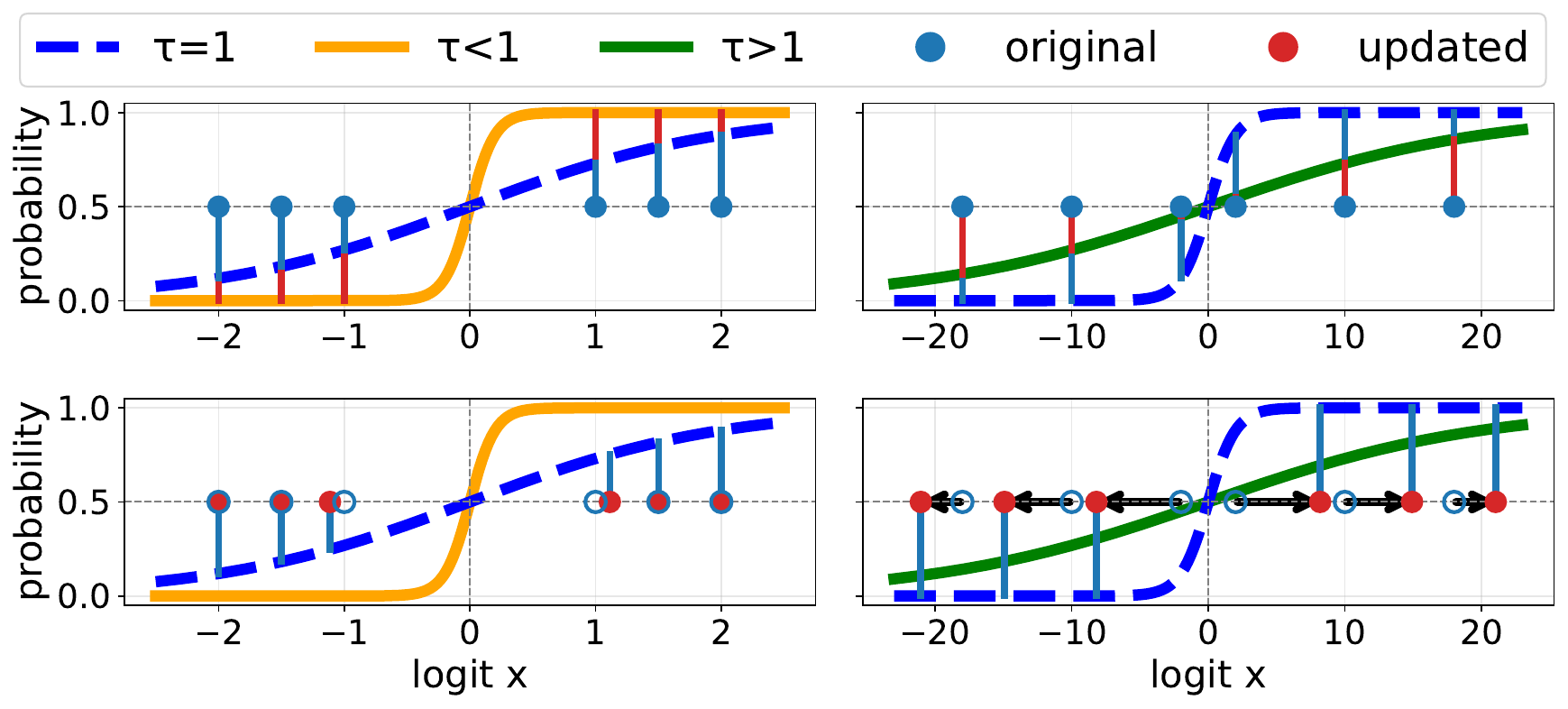}
\caption{Effect of training-time temperature scaling on logit updates and test-time confidence. Solid curves show the $\tau$-scaled sigmoid used to compute updates (\textcolor{orange}{\textbf{---}}: $\tau{<}1$, \textcolor{green}{\textbf{---}}: $\tau{>}1$), while evaluation uses the standard $\tau{=}1$ mapping (\textcolor{blue}{\textbf{-$\;$-}}), so updated logits (\textcolor{blue}{$\bullet$} $\rightarrow$ \textcolor{red}{$\bullet$}) can yield shifted probabilities and degraded calibration; note the different x-axis ranges.}
\label{fig:rationale}
\end{figure}

Figure~\ref{fig:rationale} illustrates the core mechanism of TSA in a simplified binary classification setting with a sigmoid (i.e., binary softmax) output \(\sigma(x)=\frac{1}{1+e^{-x}}\). Under standard training and evaluation, a logit \(z\) is mapped to a probability \(p=\sigma(z)\). TSA perturbs this pipeline only during training by applying temperature scaling to the probability mapping, $p_\tau \;=\; \sigma\!\left(\frac{z}{\tau}\right),$
thereby altering the geometric correspondence between logits and probabilities---and thus the gradient signal driving optimization---while evaluation still uses the unscaled mapping \(\sigma(z)\) (i.e., \(\tau=1\)).

The top row visualizes this mismatch at the level of mappings. When \(\tau<1\) (left column), the sigmoid becomes steeper and saturates earlier: logits that would be moderately confident under \(\tau=1\) can already be interpreted as high-confidence under \(\sigma(z/\tau)\). Conversely, when \(\tau>1\) (right column), the sigmoid flattens and expands the effective probability range: the same logits are mapped to less extreme probabilities, making predictions appear less confident in the \(\tau\)-scaled training space. Consequently, identical logits correspond to systematically different confidence values depending on \(\tau\), even before any parameter update is applied.

The bottom row shows how this distortion induces biased logit updates via a single gradient step. For \(\tau<1\), samples that are underconfident under the evaluation mapping \(\sigma(z)\) may already fall into a saturated, high-confidence regime under \(\sigma(z/\tau)\), where gradients are strongly attenuated; their logits therefore move little, yielding underconfident outputs at test time in the logit space. In contrast, for \(\tau>1\), samples that are already overconfident under \(\sigma(z)\) can be mapped to an underconfident regime under \(\sigma(z/\tau)\), keeping gradients ``alive'' and pushing logits further, which translates into even more overconfident predictions when evaluated with \(\tau=1\). In summary, training-time temperature scaling effectively trains the model in a distorted logit--probability space but deploys it in the original space, forcing logits toward under/over-confident regimes and inducing intended miscalibration.

However, naively applying temperature scaling during local training can induce convergence instability or degrade accuracy, since it directly reshapes the gradients that drive optimization. To address this, Section~\ref{sec:theory} provides a theoretical convergence analysis showing that TSA can preserve stable convergence under this perturbation.

\subsection{Overall Workflow}
We present the overall workflow of the temperature scaling attack in a standard federated learning setting. First, the server broadcasts the current global model to participating clients and each client then performs local training on its own data. Here, malicious clients additionally configure the local learning rate $\eta$ and the training softmax temperature $\tau$ to steer the global model toward overconfidence ($\tau{>}1$) or underconfidence ($\tau{<}1$). In particular, when an attacker configures the training temperature $\tau$, it scales the local learning rate to $\eta \cdot \tau$. We note that this learning rate-temperature ($\eta$-$\tau$) coupling is carefully chosen based on our theoretical convergence analysis (c.f., Sec.~\ref{sec:theory}). After local training, all clients, including attackers, take a typical federated learning operation and upload their model updates (e.g., parameters or gradients) to the server. Finally, the server aggregates the received updates using its aggregation rule to produce an updated global model.

\subsection{Theoretical Convergence Analysis}
\label{sec:theory}

We analyze temperature-scaled local training in the standard non-convex FL framework and characterize the role of the learning rate--temperature coupling. Consider $K$ clients collaboratively minimizing
$
F(\theta) \triangleq \sum_{k=1}^{K} p_k f_k(\theta),
$
where $\theta$ denotes the model parameters, $p_k$ is the aggregation weight, and
$f_k(\theta)=\mathbb{E}_{\xi\sim\mathcal{D}_k}[\ell(\theta;\xi)]$
is the local objective under the standard temperature $\tau=1$. In each communication round $t$, client $k$ starts from the global model $\theta_t$ and performs $E$ local SGD steps using temperature $\tau_k$ and learning rate $\eta_k$.

We adopt the standard assumptions used in non-convex FL analysis.

\begin{assumption}[Smoothness of base objective]
\label{ass:smooth_base}
For each client $k$, $f_k$ is $L$-smooth:
\begin{equation}
\|\nabla f_k(u)-\nabla f_k(v)\|
\le
L\|u-v\|,
\qquad \forall u,v.
\end{equation}
\end{assumption}

\begin{assumption}[Unbiased and bounded stochastic gradients]
\label{ass:variance}
Let $\mathbf{g}_k(\theta;\xi)$ denote a stochastic gradient of the base loss on client $k$. Then
\begin{align}
\mathbb{E}_{\xi}[\mathbf{g}_k(\theta;\xi)]
&=
\nabla f_k(\theta),\\
\mathbb{E}_{\xi}
\left[
\|\mathbf{g}_k(\theta;\xi)-\nabla f_k(\theta)\|^2
\right]
&\le
\sigma^2,
\qquad \forall\theta.
\end{align}
\end{assumption}

\begin{assumption}[Bounded heterogeneity]
\label{ass:hetero}
There exists $\Gamma^2$ such that, for all $\theta$,
\begin{equation}
\sum_{k=1}^{K}
p_k
\|\nabla f_k(\theta)-\nabla F(\theta)\|^2
\le
\Gamma^2.
\end{equation}
\end{assumption}

\begin{assumption}[Learning rate--temperature coupling]
\label{ass:beta}
Clients may use different learning rates $\eta_k$ and temperatures $\tau_k$, but satisfy
\begin{equation}
\frac{\eta_k}{\tau_k}
=
\beta,
\qquad
\forall k\in[K].
\label{eq:beta_invariance}
\end{equation}
\end{assumption}

\noindent{\textit{Effect of $\eta$--$\tau$ coupling}:}
For a sample $\xi=(x,y)$, let
$
\mathbf{p}_{\tau}
=
\mathrm{softmax}(\mathbf{z}(\theta;x)/\tau).
$
The gradient of temperature-scaled softmax cross-entropy satisfies
\begin{equation}
\tau\nabla_\theta\ell_{\tau}(\theta;\xi)
=
\nabla_\theta\ell_{1}(\theta;\xi)
+
\mathbf{r}_{\tau}(\theta;\xi),
\label{eq:gradient_decomposition}
\end{equation}
where
\begin{equation}
\mathbf{r}_{\tau}(\theta;\xi)
\triangleq
(\mathbf{J}_{\theta}\mathbf{z})^\top
(\mathbf{p}_{\tau}-\mathbf{p}_{1}).
\end{equation}
Therefore, under $\eta_k=\beta\tau_k$,
\begin{equation}
\eta_k\nabla_\theta\ell_{\tau_k}
=
\beta\nabla_\theta\ell_{1}
+
\beta\mathbf{r}_{\tau_k}.
\label{eq:coupled_gradient}
\end{equation}
Thus, the coupling cancels the explicit $1/\tau_k$ factor in the gradient, while $\mathbf{r}_{\tau_k}$ captures the remaining change caused by the temperature-dependent predictive distribution. In particular, $\mathbf{r}_{1}=0$ for benign training.

We assume that this remaining perturbation is bounded:
\begin{equation}
\mathbb{E}_{\xi\sim\mathcal{D}_k}
\left[
\|\mathbf{r}_{\tau_k}(\theta;\xi)\|^2
\right]
\le
\rho_k^2,
\quad
\forall k,\theta,
\qquad
\sum_{k=1}^{K}p_k\rho_k^2\le\rho^2.
\label{eq:residual_bound}
\end{equation}
Appendix~\ref{app:theory} provides the temperature-dependent smoothness analysis and the detailed convergence derivation.

\begin{theorem}[Non-convex convergence with variable $\tau$]
\label{thm:convergence}
Suppose Assumptions~\ref{ass:smooth_base}--\ref{ass:beta} and Eq.~\eqref{eq:residual_bound} hold, $F^\star=\inf_{\theta}F(\theta)>-\infty$, client sampling and aggregation are unbiased with respect to the global objective, and $\beta$ satisfies the stability condition given in Appendix~\ref{app:theory}. Then
\begin{equation}
\label{eq:nonconvex_beta_bound}
\begin{aligned}
\frac{1}{T}\sum_{t=0}^{T-1}
\mathbb{E}\!\left[\|\nabla F(\theta_t)\|^2\right]
\le\;&
\frac{2(F(\theta_0)-F^\star)}{\beta ET}+O(\rho^2)
\\
&+
O\!\left(
\frac{\beta L(\sigma^2+E\Gamma^2)}{M}
\right)
\\
&+
O\!\left(
\beta^2L^2E^2(\Gamma^2+\sigma^2)
\right).
\end{aligned}
\end{equation}
Here, $T$ is the number of communication rounds and $M$ is the number of participating clients per round.
\end{theorem}

The optimization, stochastic-variance, and local-drift terms in Eq.~\eqref{eq:nonconvex_beta_bound} have the standard non-convex FL form. The additional $O(\rho^2)$ term captures the deviation introduced by temperature scaling. When $\tau_k=1$ for all clients, $\rho=0$ and the bound reduces to the benign case. More generally, the $\eta$--$\tau$ coupling controls the primary update scale without eliminating the temperature-dependent residual that alters predictive confidence. When this residual remains moderate, the resulting convergence behavior can therefore remain close to benign training while TSA still induces systematic calibration shifts.

\section{Evaluation}
\label{sec:eval}
\subsection{Setup}
\label{subsec:setup}
We now present extensive experimental results using three benchmarks, including comparisons against diverse baselines and a set of case studies. We detail the evaluation setup below.

\noindent{}\textbf{Dataset and Models.}
For the overall evaluation, we consider three dataset-model configuration pairs that cover increasing task difficulty and model capacity:

\noindent{}\textbf{$\bullet$ MNIST-MLP:}
The MNIST handwritten digits are paired with a lightweight multi-layer perceptron (MLP), serving as a simple sanity-check benchmark for analyzing confidence manipulation under a controlled setting.

\noindent{}\textbf{$\bullet$ CIFAR10-CNN:}
For CIFAR-10 we use a standard convolutional neural network (CNN) to evaluate the attack on natural images, where calibration is known to be challenging.

\noindent{}\textbf{$\bullet$ CIFAR100-ResNet18:}
The CIFAR-100 data set is configured with ResNet-18 to study the attack under a larger label space and a stronger backbone, which typically amplifies calibration errors and stresses robustness to heterogeneity.

\noindent{}\textbf{Federated Learning Configuration.}
Unless otherwise specified, we adopt a standard synchronous FL setup using FedAvg. In each FL round, the server samples $5$ clients (10\%) uniformly at random from a total of $50$ clients and broadcasts the current global model. Each selected client performs $5$ steps of local SGD on its private data and returns either model parameters or model updates to the server, which aggregates them using data-size weights $p_k$. To reflect realistic statistical heterogeneity, we construct non-IID client partitions using a Dirichlet split~\cite{hsu2019measuring} with concentration parameter $\alpha$=1.0 (smaller $\alpha$ indicates higher heterogeneity), otherwise specified. We report results averaged over $5$ independently selected random seeds and evaluate both predictive performance and calibration using accuracy and calibration metrics.

\subsection{Overall Attack Performance}
\label{subsec:eval-overall}
\begin{table}[t]
\centering
\begin{small}
\begin{sc}
\begin{adjustbox}{width=\linewidth, center}
\begin{tabular}{cccccc}
\toprule
Temp. & Accuracy (\%)         & ECE (\%)             & sECE (\%)             & BrierScore           & NLL                  \\ \midrule
1     & 91.57 ± 0.23          & 3.11 ± 0.35          & 3.06 ± 0.35           & 0.13 ± 0.00          & 0.29 ± 0.00          \\ \midrule
2     & \textbf{91.46 ± 0.31} & 0.85 ± 0.40          & -0.56 ± 0.63          & 0.13 ± 0.00          & 0.29 ± 0.01          \\
3     & 91.33 ± 0.33          & 2.78 ± 0.68          & -2.74 ± 0.68          & 0.13 ± 0.01          & 0.32 ± 0.02          \\
4     & 91.19 ± 0.40          & 4.11 ± 0.72          & -4.11 ± 0.72          & 0.14 ± 0.01          & 0.37 ± 0.03          \\
5     & 91.09 ± 0.45          & \textbf{5.02 ± 0.72} & \textbf{-5.01 ± 0.73} & \textbf{0.14 ± 0.01} & \textbf{0.43 ± 0.04} \\ \midrule
0.5   & 91.46 ± 0.26          & 4.76 ± 0.34          & 4.75 ± 0.33           & 0.13 ± 0.00          & 0.30 ± 0.01          \\
0.33  & 91.43 ± 0.24          & 5.11 ± 0.36          & 5.10 ± 0.36           & 0.13 ± 0.00          & 0.30 ± 0.01          \\
0.25  & 91.41 ± 0.23          & 5.23 ± 0.38          & 5.22 ± 0.38           & 0.13 ± 0.00          & 0.30 ± 0.01          \\
0.2   & \textbf{91.40 ± 0.25} & \textbf{5.29 ± 0.39} & \textbf{5.28 ± 0.38}  & \textbf{0.14 ± 0.00} & \textbf{0.31 ± 0.01} \\
\bottomrule
\end{tabular}
\end{adjustbox}
\end{sc}
\caption{Global model accuracy and confidence metrics with varying temperature on MNIST-MLP configuration.}
\label{tab:mnist-result}
\end{small}
\end{table}
\begin{table}[t]
\centering
\begin{small}
\begin{sc}
\begin{adjustbox}{width=\linewidth, center}
\begin{tabular}{cccccccccc}
\toprule
Temp. & Accuracy (\%)         & ECE (\%)              & sECE (\%)              & BrierScore           & NLL                  \\ \midrule
1     & 68.55 ±   0.73        & 1.56 ± 0.59           & -0.84 ± 1.05           & 0.43 ± 0.01          & 0.89 ± 0.02          \\ \midrule
2     & \textbf{68.63 ± 0.86} & 6.01 ± 0.36           & -6.00 ± 0.35           & 0.43 ± 0.01          & 0.92 ± 0.02          \\
3     & 68.46 ± 0.61          & 9.08 ± 1.31           & -9.08 ± 1.31           & 0.44 ± 0.00          & 0.98 ± 0.04          \\
4     & 68.04 ± 0.47          & 11.66 ± 2.57          & -11.64 ± 2.59          & 0.46 ± 0.02          & 1.05 ± 0.09          \\
5     & 67.63 ± 0.79          & \textbf{13.46 ± 3.11} & \textbf{-13.43 ± 3.13} & \textbf{0.47 ± 0.03} & \textbf{1.12 ± 0.15} \\ \midrule
0.5   & \textbf{68.61 ± 0.77} & 3.89 ± 2.61           & 3.81 ± 2.67            & 0.43 ± 0.00          & 0.90 ± 0.01          \\
0.33  & 68.46 ± 0.72          & 5.96 ± 3.12           & 5.89 ± 3.14            & 0.43 ± 0.00          & 0.91 ± 0.01          \\
0.25  & 68.40 ± 0.76          & 6.68 ± 3.38           & 6.64 ± 3.40            & 0.44 ± 0.00          & 0.92 ± 0.01          \\
0.2   & 68.27 ± 0.68          & \textbf{7.72 ± 4.27}  & \textbf{7.72 ± 4.26}   & \textbf{0.44 ± 0.00} & \textbf{0.93 ± 0.01} \\ 
\bottomrule
\end{tabular}
\end{adjustbox}
\end{sc}
\caption{Global model accuracy and confidence metrics with varying temperature on CIFAR10-CNN configuration.}
\label{tab:cifar-result}
\end{small}
\end{table}
\begin{table}[t]
\centering
\begin{small}
\begin{sc}
\begin{adjustbox}{width=\linewidth, center}
\begin{tabular}{cccccccccc}
\toprule
Temp. & Accuracy (\%)         & ECE (\%)              & sECE (\%)              & BrierScore           & NLL                  \\ \midrule
1     & 36.24 ± 0.76          & 15.59 ± 0.77          & -15.59 ± 0.77          & 0.80 ± 0.01          & 2.71 ± 0.03          \\ \midrule
2     & 37.20 ± 0.84          & 24.22 ± 1.33          & -24.22 ± 1.33          & 0.84 ± 0.02          & 2.93 ± 0.08          \\
3     & \textbf{37.78 ± 0.61} & 30.24 ± 2.28          & -30.24 ± 2.28          & 0.88 ± 0.02          & 3.26 ± 0.15          \\
4     & 37.47 ± 0.71          & 35.39 ± 3.84          & -35.39 ± 3.84          & 0.93 ± 0.03          & 3.74 ± 0.38          \\
5     & 37.57 ± 0.56          & \textbf{38.19 ± 5.47} & \textbf{-38.18 ± 5.47} & \textbf{0.95 ± 0.05} & \textbf{4.15 ± 0.64} \\ \midrule
0.5   & \textbf{35.74 ± 0.67} & 10.41 ± 2.84          & -10.41 ± 2.84          & 0.79 ± 0.01          & 2.64 ± 0.08          \\
0.33  & 35.03 ± 0.65          & 9.31 ± 3.26           & -9.30 ± 3.27           & 0.79 ± 0.01          & 2.64 ± 0.06          \\
0.25  & 35.63 ± 0.72          & 7.93 ± 3.80           & -7.93 ± 3.80           & 0.79 ± 0.01          & 2.61 ± 0.06          \\
0.2   & 35.41 ± 0.43          & \textbf{11.62 ± 6.02} & \textbf{-2.93 ± 13.94} & \textbf{0.80 ± 0.04} & \textbf{2.69 ± 0.14} \\ 
\bottomrule
\end{tabular}
\end{adjustbox}
\end{sc}
\caption{Global model accuracy and confidence metrics with varying temperature on CIFAR100-ResNet18 configuration.}
\label{tab:cifar100-result}
\end{small}
\end{table}

\noindent{}\textbf{MNIST--MLP.}
On MNIST--MLP, accuracy stays nearly constant (around $91\%$) across all $\tau$, but calibration shifts in a predictable direction.
For $\tau>1$, sECE becomes increasingly negative, indicating growing \emph{over-confidence}; for $\tau<1$, sECE becomes strongly positive, indicating \emph{under-confidence}, with ECE generally increasing as $|\tau-1|$ grows.
While ECE is numerically smaller at $\tau{=}2,3$ than at $\tau{=}1$, this mainly reflects a \emph{trend shift} from a slightly under-confident benign model (positive sECE) toward over-confidence (negative sECE), not a guaranteed improvement in reliability.
At larger temperatures, calibration degrades in both signed and unsigned terms (e.g., up to $\mathrm{ECE}=5.02$ and $\mathrm{sECE}=-5.01$ at $\tau{=}5$), and proper scoring rules (Brier/NLL) also worsen.

\noindent{}\textbf{CIFAR10--CNN.}
A similar but amplified pattern holds on CIFAR10--CNN (Table~\ref{tab:cifar-result}).
Accuracy remains tightly clustered around $68\%$, yet calibration deteriorates sharply as $\tau$ moves away from $1$.
For $\tau>1$, the model becomes progressively \emph{over-confident}: ECE rises from $1.56$ ($\tau{=}1$) to $6.01, 9.08, 11.66,$ and $13.46$ for $\tau{=}2$--$5$, with sECE tracking the same direction ($-6.00$ to $-13.43$).
For $\tau<1$, the trend flips to \emph{under-confidence}: ECE increases to $3.89$--$7.72$ for $\tau{=}0.5$--$0.2$ and sECE becomes positive with comparable magnitudes ($3.81$--$7.72$).
Brier and NLL also increase as confidence distortion strengthens (e.g., NLL $0.89\!\rightarrow\!1.12$ and Brier $0.43\!\rightarrow\!0.47$ from $\tau{=}1$ to $\tau{=}5$), confirming that probability quality worsens even when accuracy is preserved.
\begin{figure*}[t!]
    \centering
    \includegraphics[width=.9\linewidth]{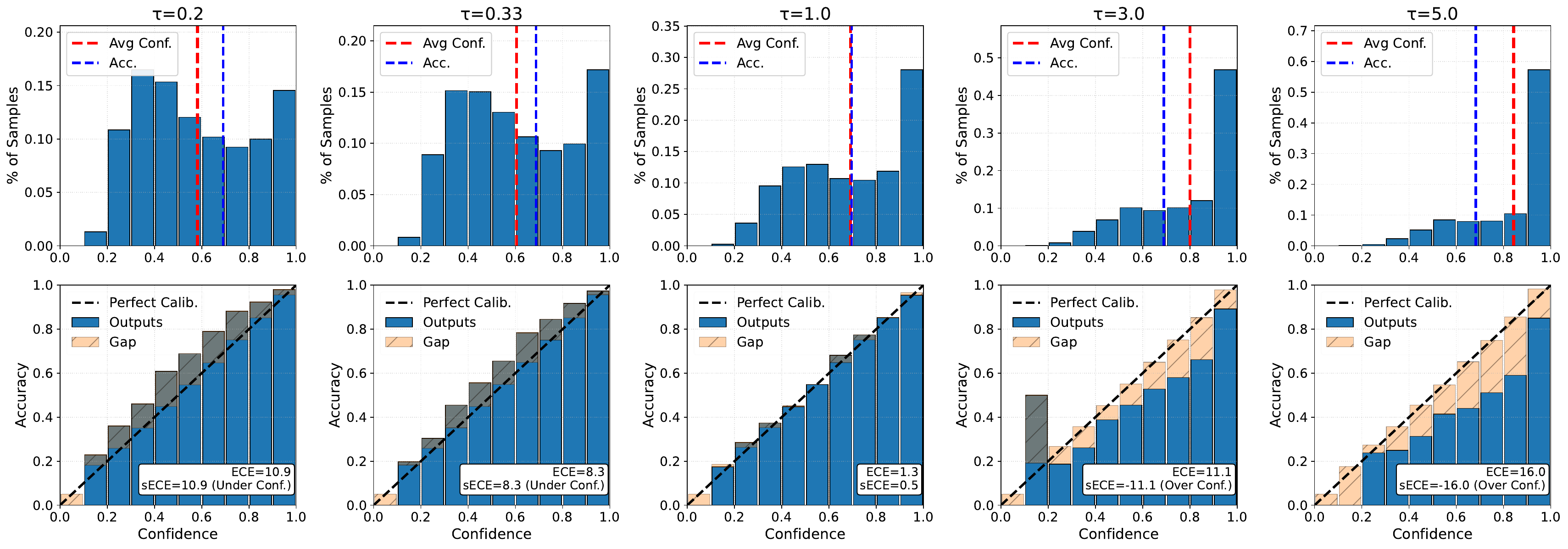}
    \caption{Confidence distribution and reliability diagrams under different training temperatures $\tau$ on CIFAR10--CNN. \textbf{Top row}: histogram of predicted confidences with \textcolor{red}{\textbf{average confidence}} (red dashed) and \textcolor{blue}{\textbf{accuracy}} (blue dashed). \textbf{Bottom row}: reliability diagrams showing the gap between bin accuracy and confidence. Low temperatures ($\tau{<}1$) induce under-confidence, while high temperatures ($\tau{>}1$) induce over-confidence, despite accuracy remaining largely unchanged.}
    \label{fig:reliability-diagram}
\end{figure*}

\noindent{}\textbf{CIFAR100--ResNet18.}
CIFAR100--ResNet18 is already severely miscalibrated and \emph{over-confident} at $\tau{=}1$ (ECE/sECE $=15.59$/$-15.59$).
Raising $\tau$ worsens calibration dramatically while accuracy stays in the mid-$30\%$ range: ECE increases to $24.22, 30.24, 35.39,$ and $38.19$ for $\tau{=}2$--$5$ (with sECE matching), and Brier/NLL deteriorate accordingly (Brier $0.80\!\rightarrow\!0.95$, NLL $2.71\!\rightarrow\!4.15$ at $\tau{=}5$).
Lowering $\tau$ can reduce ECE in some cases (e.g., $10.41$ at $\tau{=}0.5$, $7.93$ at $\tau{=}0.25$), but the behavior is less monotonic and becomes unstable at very low $\tau$ (e.g., $\tau{=}0.2$).
Overall, across all three benchmarks, temperature scaling enables large, directional calibration shifts with only minor accuracy changes.

\noindent{}\textbf{Calibration drift under temperature scaling.}
Figure~\ref{fig:reliability-diagram} visualizes how training-time temperature scaling systematically distorts the global model's confidence behavior on CIFAR10--CNN.
When $\tau{<}1$, the confidence mass shifts toward smaller values and the reliability curve lies above the diagonal, indicating under-confidence where empirical accuracies exceed predicted confidences across bins.
When $\tau{>}1$, confidences concentrate near one, and the reliability curve falls below the diagonal, revealing a pronounced over-confidence gap where predicted confidences consistently exceed empirical accuracies.
Notably, these calibration shifts occur while accuracy remains largely stable, confirming that temperature scaling provides a direct control knob for steering the \emph{direction} of miscalibration without requiring large changes in accuracy.

\subsection{Update and Representation Analysis}
\begin{figure}[t!]
    \centering
    \includegraphics[width=.9\linewidth]{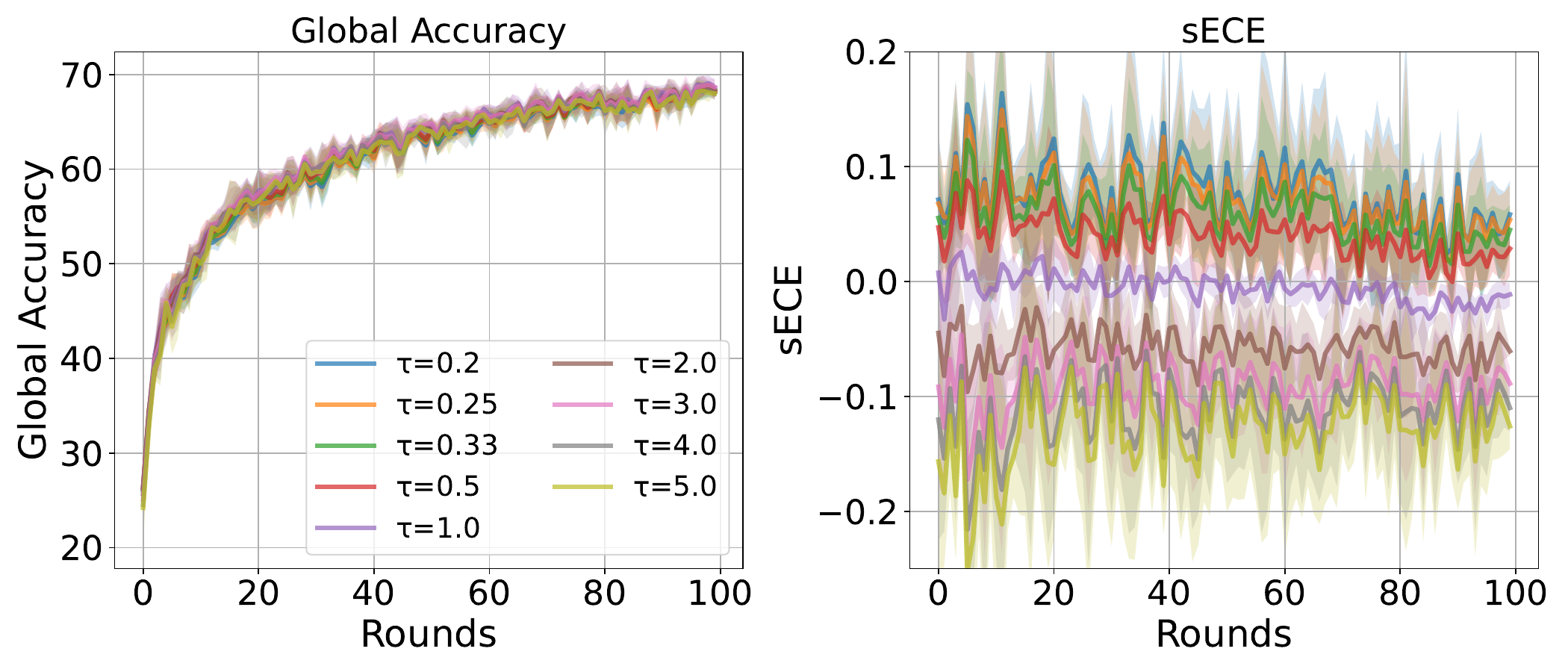}
    \caption{Global accuracy (left) and sECE (right) over rounds on CIFAR10--CNN for different training temperatures $\tau$. Accuracy remains similar across $\tau$, while sECE shifts systematically, indicating controllable under-/over-confidence.}
    \label{fig:acc-sece}
\end{figure}
\label{subsec:detectability}
In this section, we characterize how TSA affects signals commonly used to monitor federated training. We examine three complementary views: (i) the global convergence trajectory, (ii) client-update geometry, and (iii) learned representations. This analysis is intended to understand how calibration manipulation differs from conventional poisoning in signals observed by existing monitoring mechanisms.


\noindent{}\textbf{Convergence Trajectory.}
A straightforward server-side defense is to monitor the convergence trajectory using a held-out validation set, e.g., by tracking round-wise accuracy or loss. Figure~\ref{fig:acc-sece} plots the global test accuracy (left) and sECE (right) over communication rounds on CIFAR10--CNN. As shown, the accuracy trajectories under different training temperatures almost completely overlap, with no statistically meaningful separation, indicating that accuracy-only monitoring is insufficient to reveal the presence of the attack. In contrast, sECE shifts systematically with temperature: for $\tau{>}1$, sECE becomes more negative than the benign setting ($\tau{=}1$), reflecting increased over-confidence, whereas for $\tau{<}1$, sECE moves to positive values, indicating under-confidence. These results suggest that observing standard convergence metrics alone is unlikely to detect temperature scaling attacks, even though the attack induces pronounced and directional calibration drift.

\begin{figure}[t!]
    \centering
    \includegraphics[width=.9\linewidth]{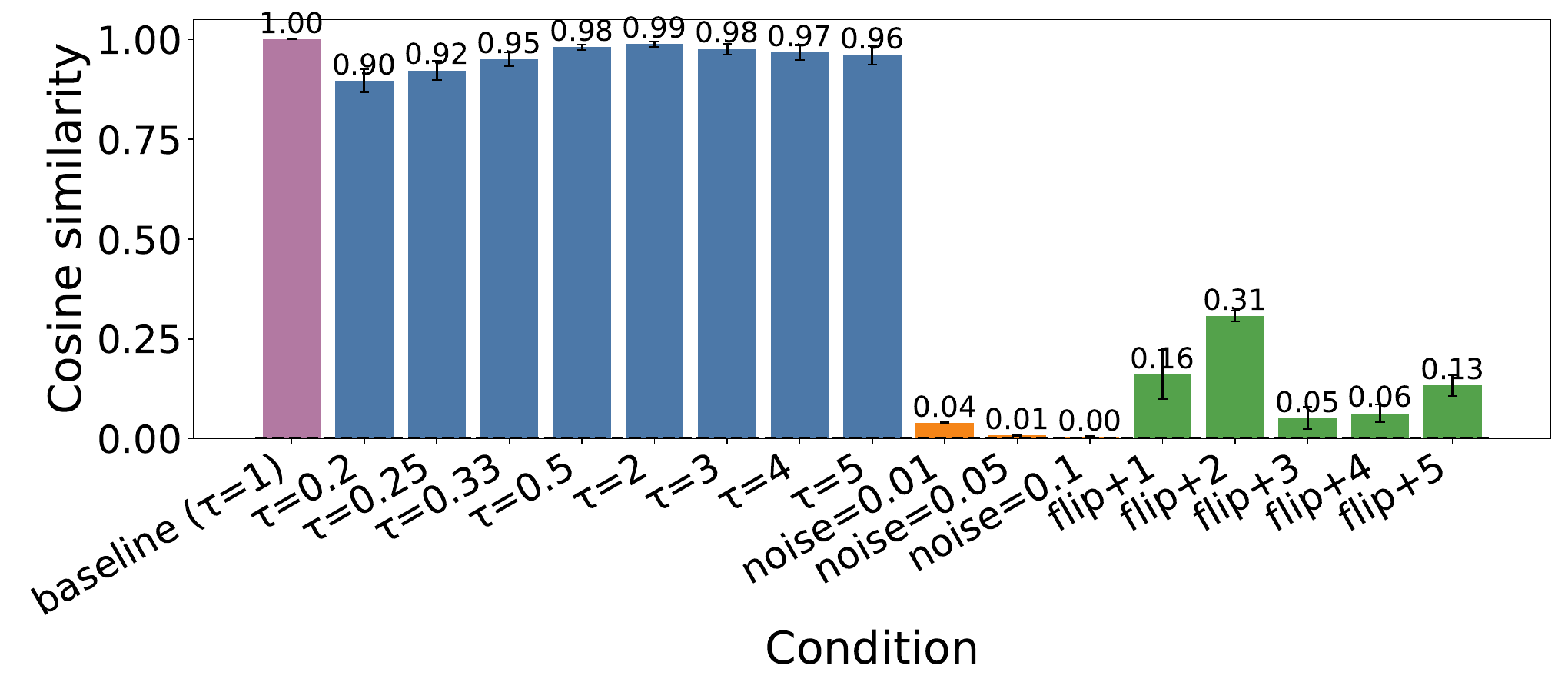}
    \caption{Cosine similarity between benign updates (baseline, $\tau{=}1$) and updates under different attacks. Temperature scaling maintains high similarity with benign updates, whereas noise addition and label flipping sharply reduce similarity.}
    \label{fig:lr-cosim-1}
\end{figure}

\noindent{}\textbf{Parameter Space Analysis.}
Beyond simple trajectory monitoring, many existing defenses in federated learning analyze client-submitted parameters or gradients, leveraging the intuition that malicious clients will produce \emph{atypical} updates in parameter space.
To evaluate temperature scaling attack under such update-space scrutiny, we measure update-direction similarity after $10$ local SGD steps on CIFAR10--CNN. Specifically, we define the local update as the parameter displacement $\Delta \theta \triangleq \theta_{\text{after}}-\theta_{\text{before}}$ and compute the cosine similarity $\cos(\Delta \theta_{\tau=1}, \Delta \theta_{\text{attack}})$, where $\Delta \theta_{\tau=1}$ is the benign update obtained with $\tau{=}1$ and $\Delta \theta_{\text{attack}}$ is the update produced under a given attack condition.
To contextualize the results, we compare against two conventional poisoning baselines: noise injection and label flipping.
For noise injection, we perturb the update as $\Delta \theta \leftarrow \Delta \theta + \epsilon$ with $\epsilon\sim\mathcal{N}(0,\sigma^2)$ and $\sigma\in\{10^{-1},5\times10^{-2},1\times10^{-2}\}$.
For label flipping, we shift labels by $k$ with modulo over $C$ classes, $y\leftarrow(y+k)\bmod C$, using $k\in\{1,2,3,4,5\}$ (larger shifts are redundant by symmetry).
Figure~\ref{fig:lr-cosim-1} visualizes the cosine similarity between benign updates and the updates produced by each attack.

As shown in Figure~\ref{fig:lr-cosim-1}, conventional poisoning attacks substantially reduce cosine similarity, typically around $0.2$ or lower, indicating that their updates deviate markedly from benign optimization directions. In contrast, temperature scaling preserves near-collinearity with benign updates across a wide range of temperatures, with the minimum similarity still around $0.93$. This suggests that temperature scaling attacks can remain difficult to flag using parameter space analysis, even while inducing systematic and directional confidence manipulation.

\begin{figure}[t!]
    \centering
    \includegraphics[width=.85\linewidth]{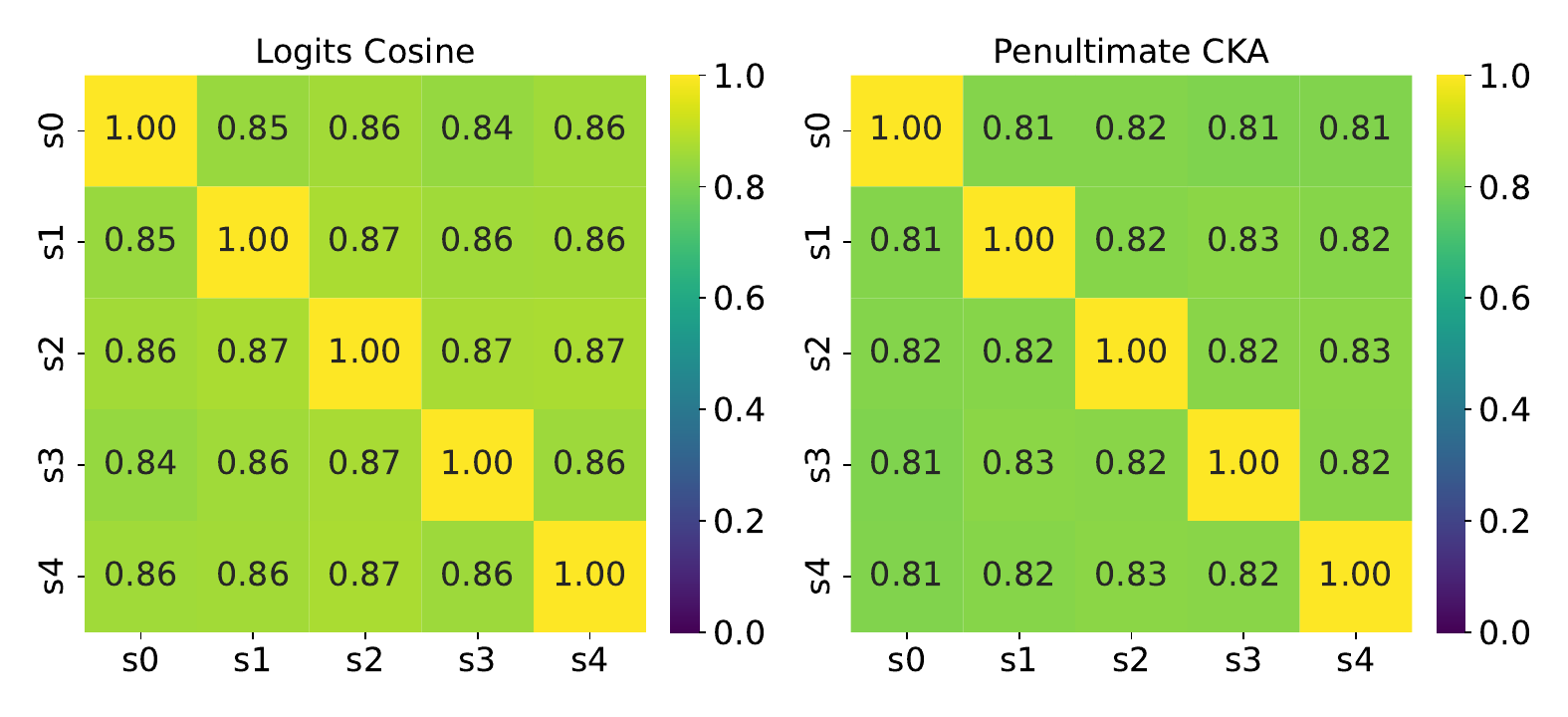}
    \caption{Seed-to-seed similarity under benign training ($\tau{=}1$): pairwise logit cosine similarity (left) and penultimate-layer CKA (right) on the same testset.}
    \label{fig:seed-variability}
\end{figure}

\noindent{}\textbf{Representation Space Analysis.}
We next study whether representation-based monitoring, which is orthogonal to trajectory- and parameter-space checks, can distinguish temperature scaling attacks.
We focus on two commonly used signals: (i) the penultimate-layer representation, which reflects the high-dimensional features learned by the network, and (ii) the classifier logits, which capture the model's categorical response to an input.
To compare penultimate representations, we use Centered Kernel Alignment (CKA), which is invariant to common symmetries such as orthogonal rotations and is thus better suited than cosine similarity for high-dimensional feature maps.
For logits, we use cosine similarity.
Figure~\ref{fig:seed-variability} first establishes a benign reference by training the same CIFAR10--CNN configuration with five different random seeds (5 epochs) and computing pairwise similarities.
Even without any attack, independent runs exhibit non-trivial variability, yet both logit cosine similarity and penultimate CKA typically remain above $\sim$0.8.
We treat this range as a practical baseline for ``benign'' similarity.

\begin{figure}[t!]
    \centering
    \includegraphics[width=.9\linewidth]{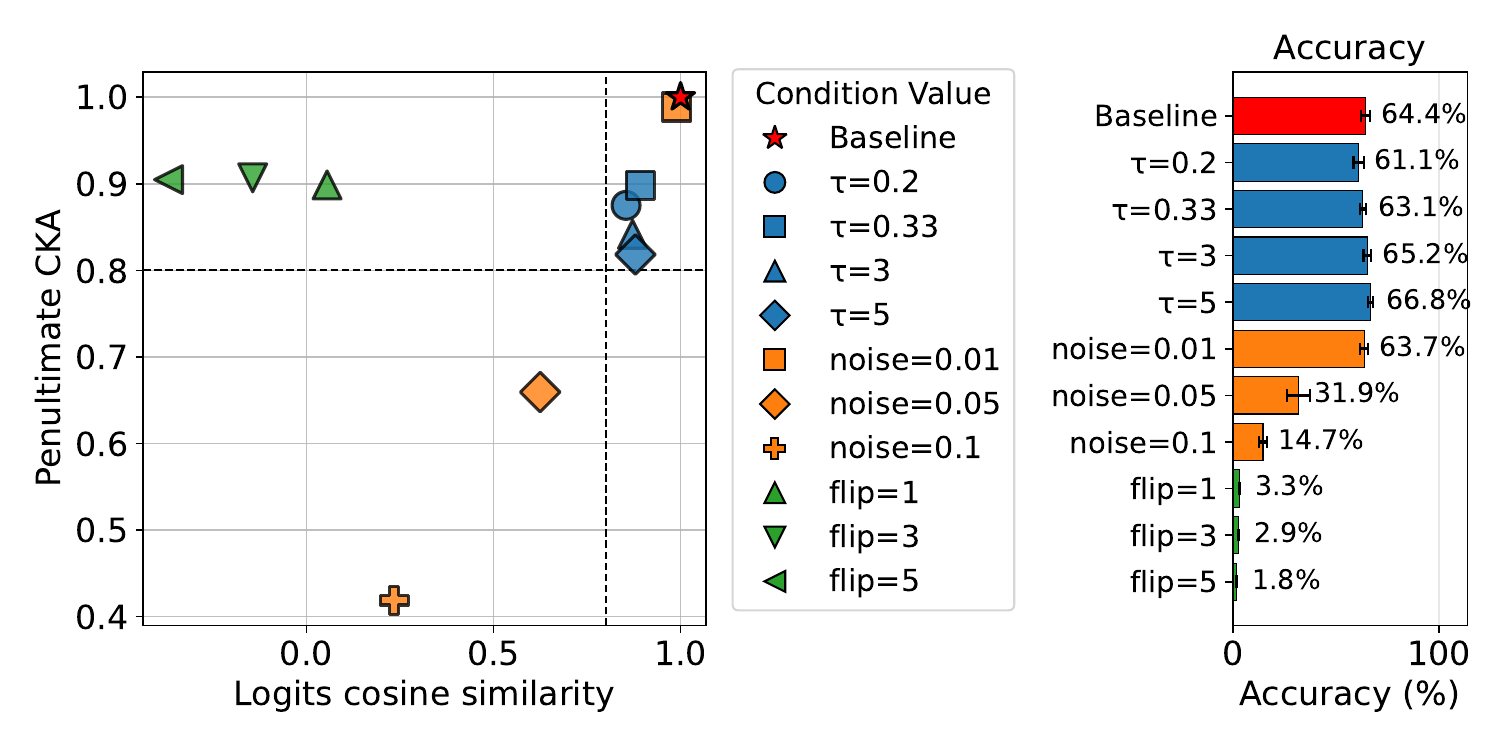}
    \caption{Logit cosine similarity vs.\ penultimate-layer CKA (left) and test accuracy (right) under different attack conditions, measured against the benign baseline.}
    \label{fig:rep_logit_similarity_scatter}
\end{figure}

Using this reference, Figure~\ref{fig:rep_logit_similarity_scatter} compares attacked models against the benign baseline.
Conventional poisoning baselines produce readily observable deviations.
Noise injection exhibits a clear trade-off: the mild setting ($\sigma{=}0.01$) stays relatively close to the baseline in similarity but also fails to meaningfully impact accuracy, whereas larger noise levels ($\sigma{=}0.05,0.1$) sharply reduce similarity and severely degrade accuracy.
Label flipping shows a distinctive signature: penultimate CKA remains high while logit cosine similarity becomes very low, consistent with preserving feature extractors but permuting output semantics.

In contrast, temperature scaling remains largely within the benign similarity range while preserving utility.
Across all tested temperatures, logit cosine similarity stays high ($\ge 0.85$) and penultimate CKA remains above $\sim$0.8, with accuracy comparable to (and sometimes slightly higher than) the baseline (e.g., $61.1\sim66.8\%$ vs.\ $64.4\%$).
Overall, these results show that the $\eta$--$\tau$ coupling substantially reduces the update-direction change introduced by temperature scaling, allowing calibration manipulation without the pronounced optimization drift observed when the coupling is removed.

\begin{figure}[t!]
    \centering
    \includegraphics[width=0.9\linewidth]{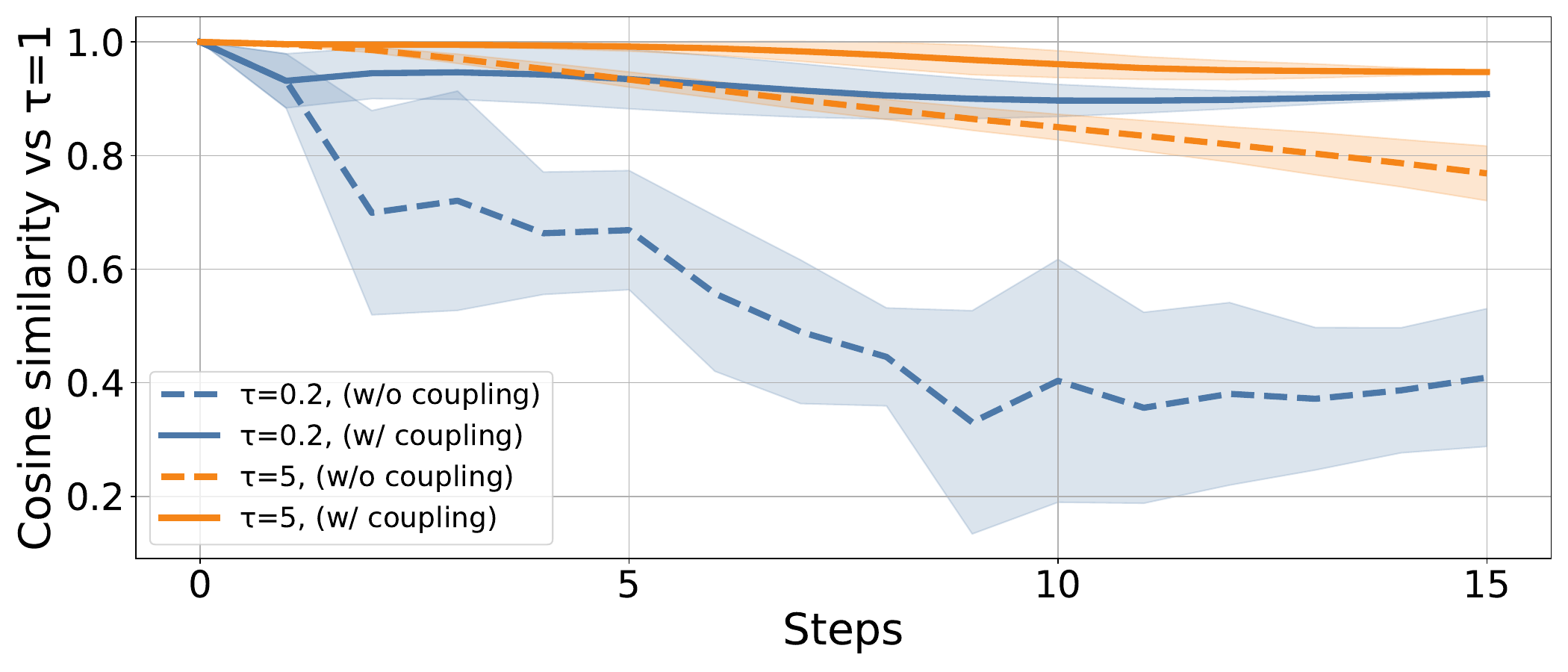}
    \caption{Cosine similarity to the $\tau{=}1$ update across local steps for $\tau{=}0.2$ and $\tau{=}5$, with and without $\eta$--$\tau$ coupling ($\beta=\eta/\tau$). Coupling preserves high similarity over steps, while removing it causes updates to diverge.}
    \label{fig:lr-cosim-2}
\end{figure}

\noindent{}\textbf{Role of $\eta$--$\tau$ coupling.}
Importantly, the benign-like behavior of the temperature scaling attack is not incidental.
We attribute this property to the proposed effective step-size ($\beta$) control, i.e., the learning rate--temperature coupling that enforces a constant $\beta=\eta/\tau$.
To isolate its impact, Figure~\ref{fig:lr-cosim-2} measures the cosine similarity to the benign update across increasing numbers of local SGD steps, comparing temperature scaling with and without $\eta$--$\tau$ coupling for $\tau{=}0.2$ and $\tau{=}5$.

With coupling, updates remain highly aligned with the benign direction throughout local training in both cases. In contrast, without coupling, changing $\tau$ directly perturbs the optimization dynamics, and the update direction drifts away from the benign trajectory as local steps accumulate. We observe a more dramatic drop for $\tau{=}0.2$ than for $\tau{=}5$, which is consistent with prior findings that $\tau{<}1$ can induce more aggressive training dynamics than $\tau{>}1$ during optimization~\cite{lee2026improving}. Overall, these results confirm that $\eta$--$\tau$ coupling is crucial for preserving benign-like update geometry, thereby enabling calibration manipulation while keeping the attack difficult to flag.
\begin{figure}
    \centering
    \includegraphics[width=0.9\linewidth]{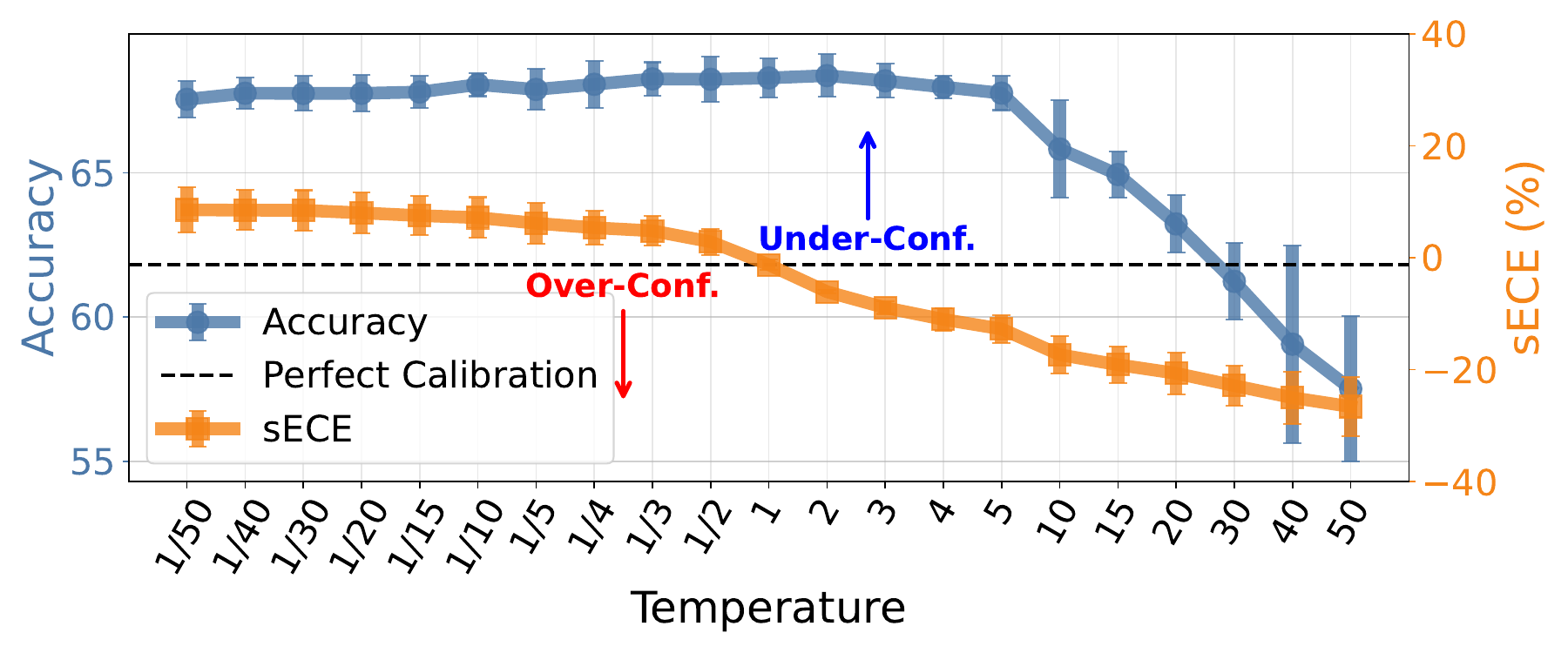}
    \caption{Effect of extreme temperatures under a fixed effective step size $\beta=\eta/\tau$ on CIFAR10--CNN. While sECE decreases monotonically as $\tau$ increases, extreme temperatures make the temperature-scaled update deviate further from its benign counterpart, causing a noticeable accuracy drop beyond $\tau{\approx}10$ and violating the accuracy-preservation goal of the attack.}
    \label{fig:temp-sweep}
\end{figure}

\noindent{}\textbf{How far can $\tau$ be pushed under a fixed $\beta$?}
The preceding analysis shows that enforcing a constant effective step size $\beta=\eta/\tau$ is critical for producing benign-like updates.
A natural question is whether the attacker can choose arbitrarily small or large temperatures as long as the $\eta{-}\tau$coupling is satisfied.
Figure~\ref{fig:temp-sweep} sweeps $\tau$ while keeping $\beta$ fixed and tracks both global accuracy and sECE. As $\tau$ increases, sECE decreases monotonically, showing that the attacker can continuously steer the model toward over-confidence.
However, beyond a moderate range, overly large $\tau$ noticeably degrades accuracy (starting around $\tau\approx 10$), violating the attack goal of preserving predictive performance.

This breakdown is consistent with the residual view in Section~\ref{sec:theory}. While fixing $\beta$ removes the explicit temperature-dependent gradient scale, it does not eliminate the change in the predictive distribution.
At high temperatures, $\mathbf{p}_\tau=\mathrm{softmax}(\mathbf{z}/\tau)$ approaches a near-uniform distribution and can deviate substantially from $\mathbf{p}_1$, increasing the residual $\mathbf{r}_\tau$ in Eq.~\eqref{eq:gradient_decomposition}.
The resulting update can therefore move farther from its benign counterpart and eventually degrade predictive accuracy.
At the other extreme, very small $\tau$ can likewise increase the residual and also tighten the step-size stability condition, as shown by the temperature-dependent smoothness analysis in the Appendix.
Thus, $\eta{-}\tau$ coupling controls the primary update scale but does not guarantee benign-like optimization for arbitrary temperatures. Achieving a favorable impact--stealth trade-off therefore requires choosing $\tau$ within a moderate range.

\begin{figure}[t!]
    \centering
    \includegraphics[width=\linewidth]{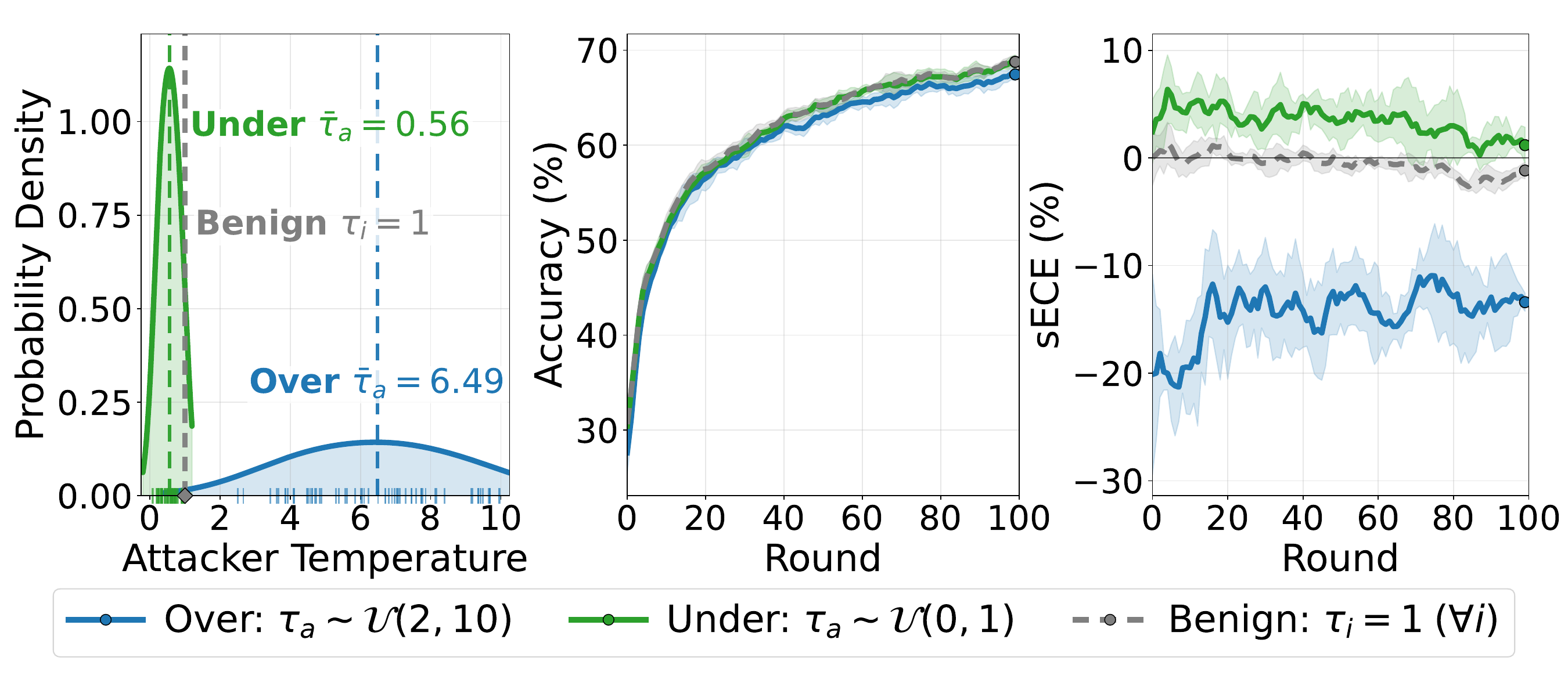}
    \caption{Attack performance with heterogeneous attacker temperatures. Attackers independently sample $\tau$ from $\mathcal{U}(0,1)$ for under-confidence and $\mathcal{U}(2,10)$ for over-confidence. Accuracy remains similar, while sECE shifts consistently in the intended direction.}
    \label{fig:random-temp}
\end{figure}

\noindent{}\textbf{Should attackers use the same $\tau$?}
The previous experiments assume that all malicious clients use the same temperature. In practice, however, coordinating an identical $\tau$ across attackers may be unrealistic. Figure~\ref{fig:random-temp} evaluates a heterogeneous setting in which each attacker independently samples its training temperature. For the under-confidence attack, we sample $\tau_a\sim\mathcal{U}(0,1)$, whereas for the over-confidence attack, we sample $\tau_a\sim\mathcal{U}(2,10)$. The resulting average attacker temperatures are $0.56$ and $6.49$, respectively.

As shown, heterogeneous temperatures produce accuracy trajectories similar to benign training, while sECE remains clearly shifted in the intended direction. This indicates that TSA does not require all attackers to share an identical temperature; maintaining the $\eta$--$\tau$ coupling for each attacker is sufficient to preserve stable optimization while collectively steering calibration. However, attackers should still agree on the \emph{direction} of the attack. If some attackers target under-confidence while others target over-confidence, their effects can partially cancel during aggregation. Thus, TSA tolerates heterogeneous temperature values, but requires directional consistency among malicious clients.

\begin{figure}[t!]
    \centering
    \includegraphics[width=.9\linewidth]{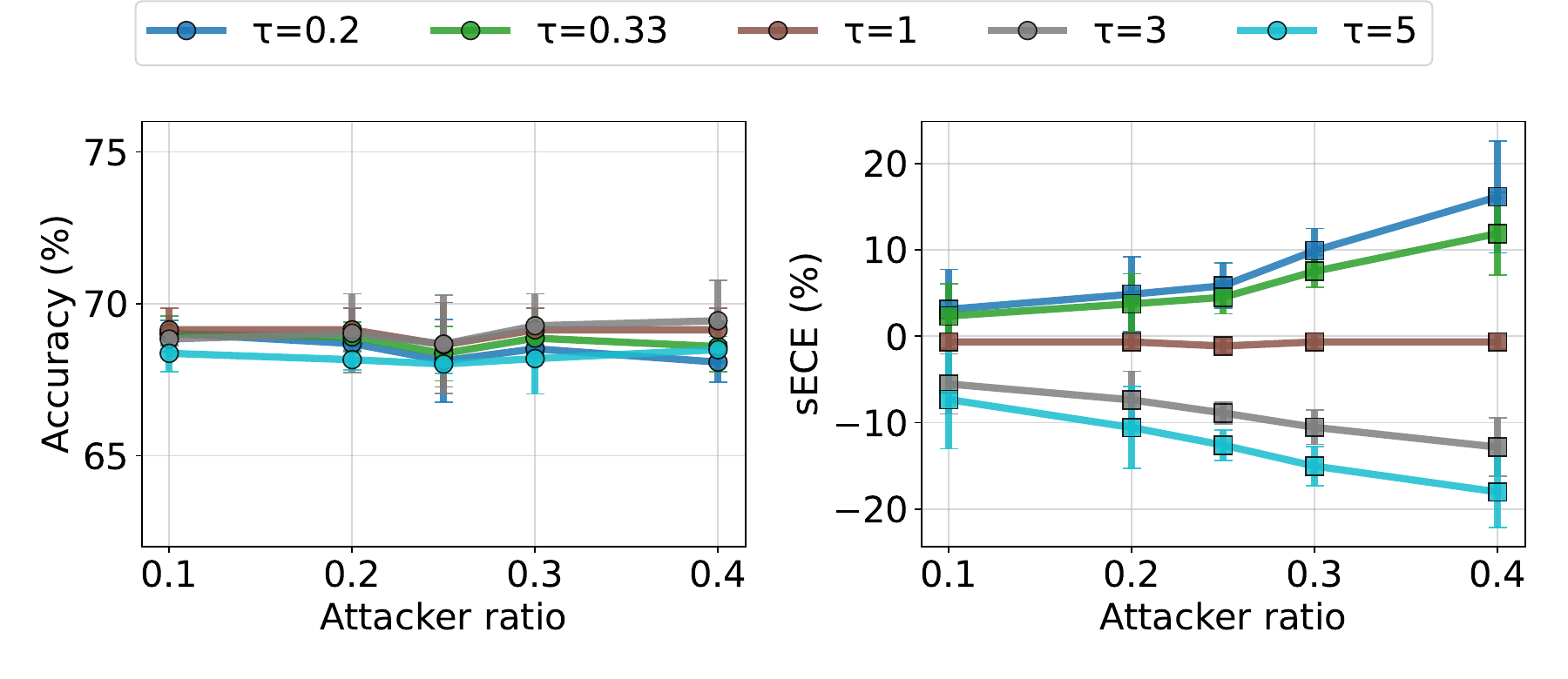}
    \caption{Effect of attacker ratio on global accuracy (left) and sECE (right) under different training temperatures $\tau$.}
    \label{fig:atk-ratio-acc-sece}
\end{figure}
\begin{figure}[t!]
    \centering
    \includegraphics[width=.95\linewidth]{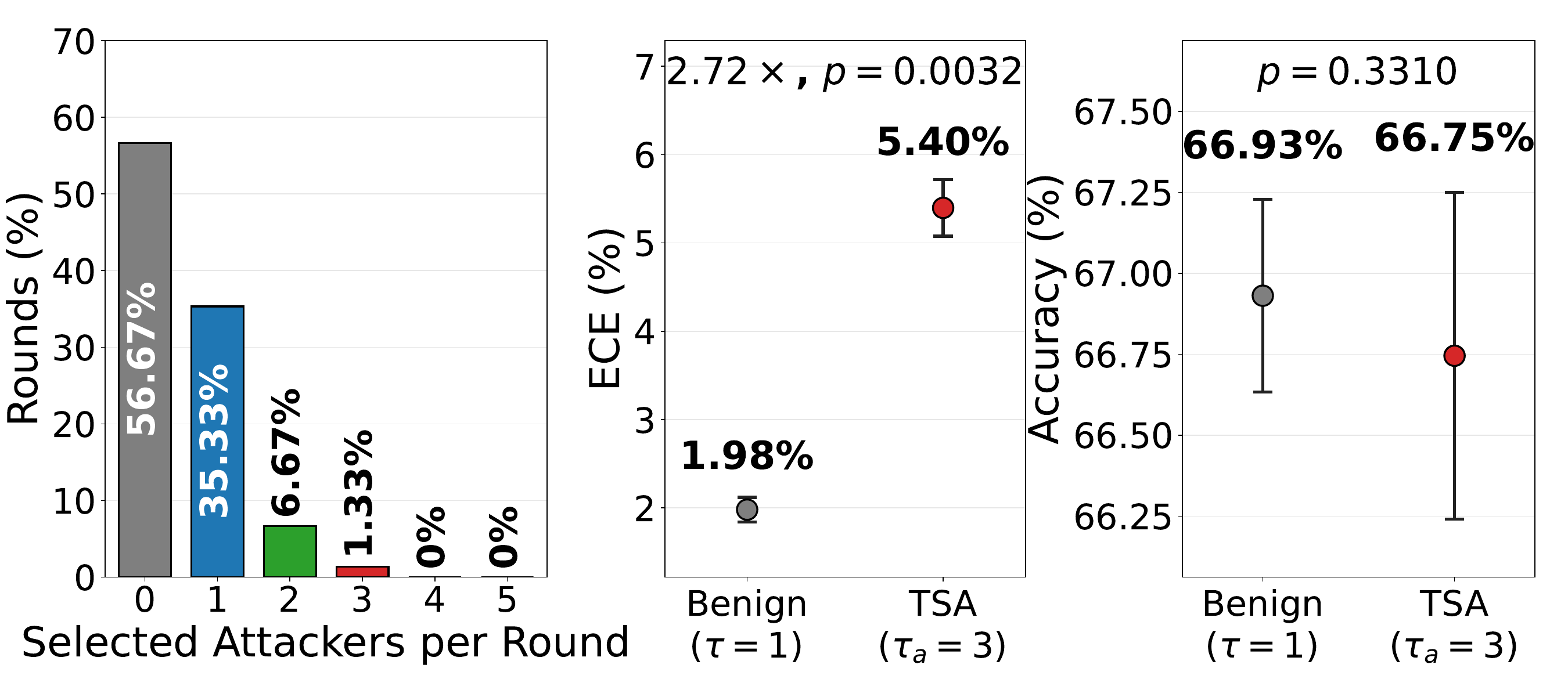}
    \caption{Impact of sporadic attacker participation.}
    \label{fig:sporadic}
\end{figure}

\subsection{Robustness and Sensitivity Analysis}
\label{subsec:sensitivity-analysis}

We now conduct a sensitivity analysis over three key factors that jointly govern the effectiveness and practicality of the proposed temperature scaling attack in federated learning: (i) the attacker ratio, (ii) the degree of non-IID data heterogeneity, and (iii) the effective step size $\beta$ (i.e., $\eta$--$\tau$ coupling).

\noindent{}\textbf{Resilience to Malicious Participation (Attacker ratio).}
Figure~\ref{fig:atk-ratio-acc-sece} reports how the global model's test accuracy (left) and signed ECE (right) change as we vary the attacker ratio in the client population.
Across all temperatures, accuracy stays nearly flat as the attacker ratio increases from $0.1$ to $0.4$, indicating that the attack remains hard to notice through standard utility monitoring.
In contrast, calibration drift amplifies with more attackers.
For high-temperature settings ($\tau>1$), sECE becomes increasingly negative as the attacker ratio grows, reflecting stronger \emph{over-confidence}; for low temperatures ($\tau<1$), sECE becomes increasingly positive, reflecting stronger \emph{under-confidence}.
Notably, even with only $10\%$ attackers, sECE already separates from the benign reference ($\tau{=}1$), and at $40\%$ attackers the deviation reaches roughly $\pm 18$--$19$ percentage points depending on the attack direction.
Overall, this result shows that attack impact scales with attacker participation while largely preserving accuracy, reinforcing that calibration is a more sensitive signal than performance for auditing this threat.

\noindent{}\textbf{Impact of sporadic attacker participation.}
We further examine whether TSA requires malicious clients to participate in every communication round.
We consider a $10\%$ attacker population (5 malicious clients among 50 total clients), with five clients randomly sampled per round.
Across three independent runs of 100 rounds each, no attacker is selected in $56.67\%$ of the rounds, while 1, 2, and 3 attackers are selected in $35.33\%$, $6.67\%$, and $1.33\%$ of the rounds, respectively.
Notably, across all three runs, we never observe a round in which four or more attackers are simultaneously selected.
Thus, malicious participation is highly sporadic: attackers are absent in more than half of the rounds, and rounds containing multiple attackers are rare.

Nevertheless, as shown in Figure~\ref{fig:sporadic}, TSA with $\tau_a{=}3$ increases ECE from $1.98\%$ to $5.40\%$ ($2.72\times$, $p{=}0.0032$).
For each seed, ECE and accuracy are first averaged over rounds 51--100 and then averaged across the three seeds; the reported $p$-values are obtained using paired $t$-tests over these seed-level averages.
Accuracy changes only marginally from $66.93\%$ to $66.75\%$, with no statistically significant difference ($p{=}0.3310$).
These results show that substantial calibration degradation can accumulate even when malicious clients participate only intermittently, while no statistically significant change in predictive accuracy is observed.

\begin{figure}[t!]
    \centering
    \includegraphics[width=.9\linewidth]{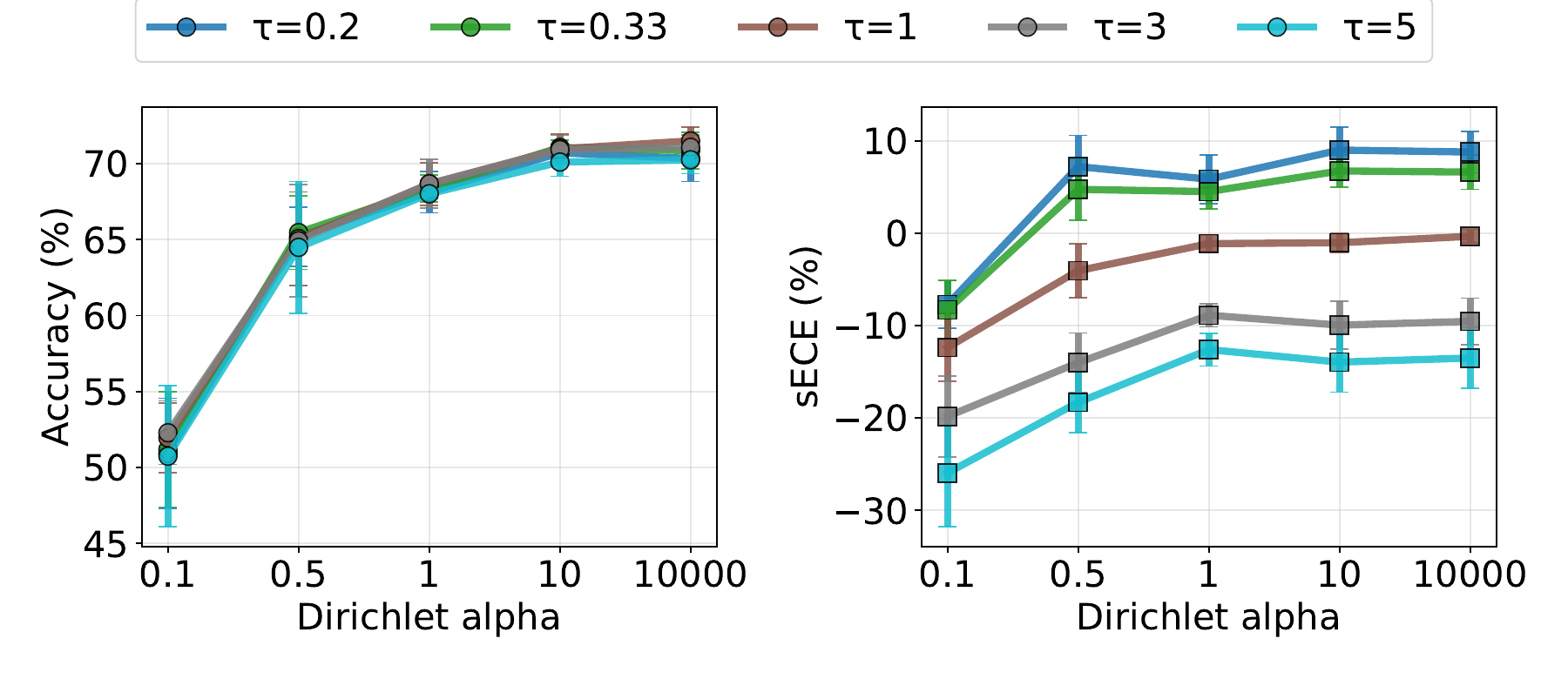}
    \caption{Effect of non-IID degree (Dirichlet $\alpha$) on global accuracy and sECE under different training temperatures $\tau$.}
    \label{fig:alpha-acc-sece}
\end{figure}

\noindent{}\textbf{Robustness to Statistical Heterogeneity (Non-IID).}
Non-IID data distributions are the norm in practical federated learning, making it essential to evaluate whether temperature scaling remains effective under heterogeneous client data.
Figure~\ref{fig:alpha-acc-sece} varies the Dirichlet concentration parameter $\alpha$ to control the strength of non-IIDness and reports the resulting global accuracy and sECE.
As expected, stronger heterogeneity (smaller $\alpha$, e.g., $\alpha{=}0.1$) degrades overall utility, and the calibration baseline also shifts (sECE becomes more biased), reflecting the increased training difficulty under severe distribution skew.
Crucially, however, the \emph{relative} effect of temperature scaling remains consistent across $\alpha$:
for $\tau<1$, sECE systematically increases (under-confidence), while for $\tau>1$, sECE decreases (over-confidence),
with accuracy trajectories remaining tightly clustered across temperatures within each $\alpha$ setting.
This indicates that the attack is robust to realistic non-IID conditions, even as heterogeneity changes the absolute operating point, the attacker can still steer the \emph{direction} of miscalibration in a controlled way without introducing conspicuous accuracy drops.

\begin{figure}[t!]
    \centering
    \includegraphics[width=.9\linewidth]{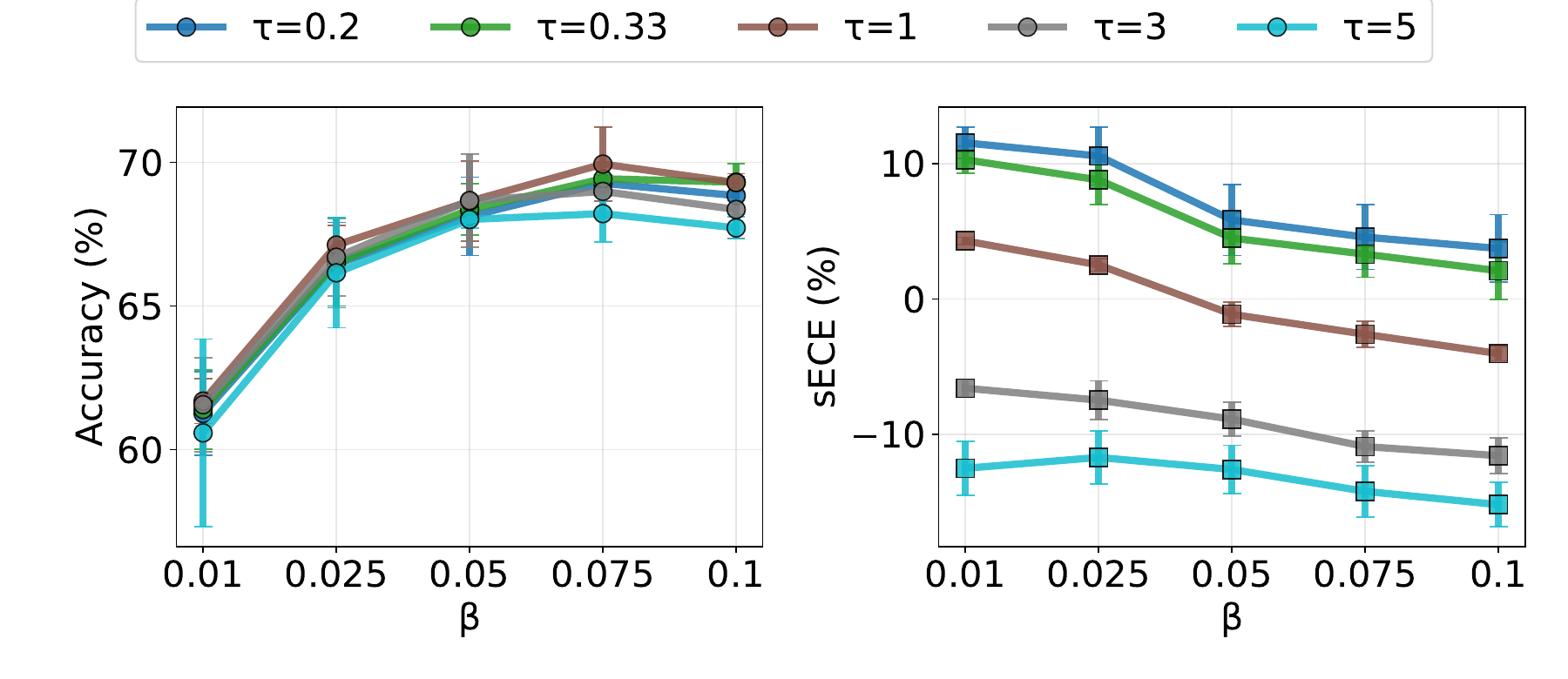}
    \caption{Sensitivity to the effective step size $\beta$ under different training temperatures $\tau$.}
    \label{fig:beta-acc-sece}
\end{figure}
\begin{figure}[t!]
    \centering
    \includegraphics[width=\linewidth]{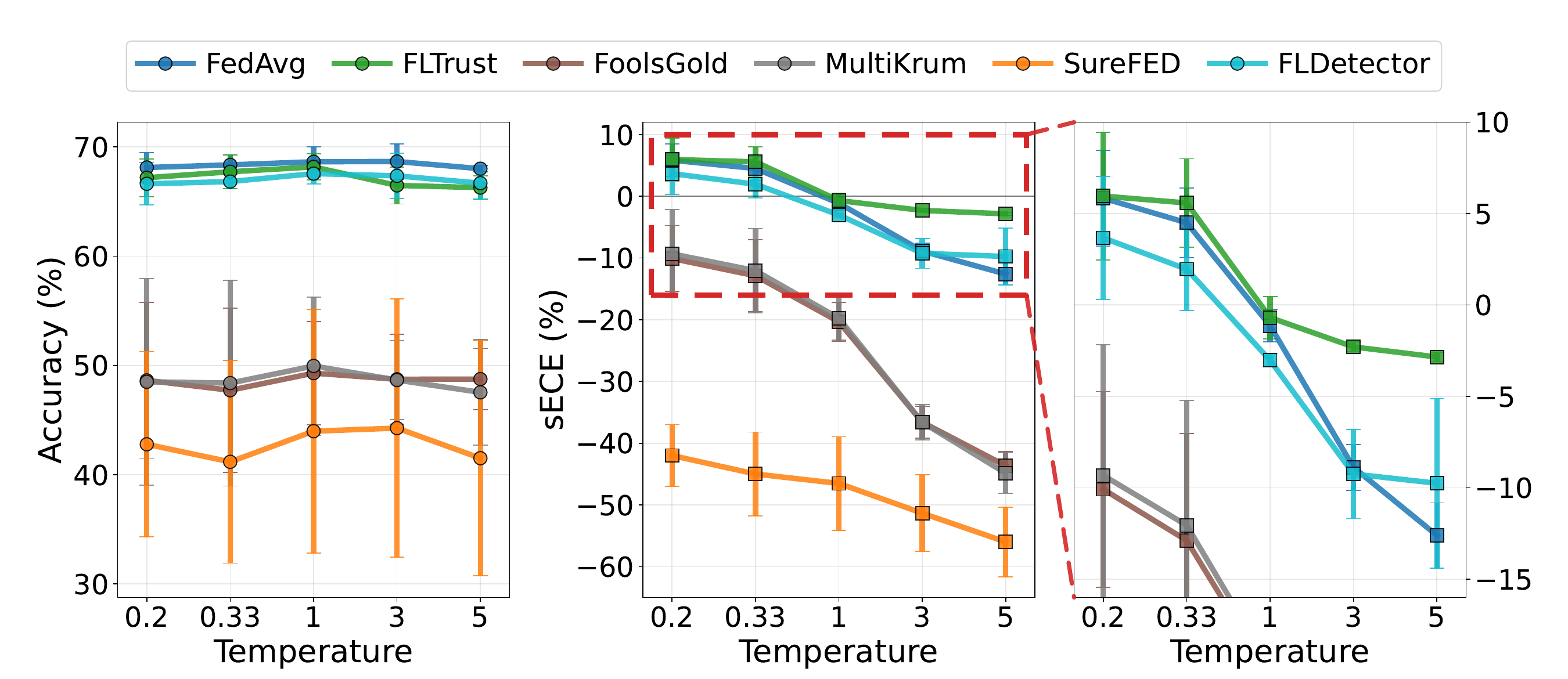}
    \caption{Accuracy and sECE under different aggregation defenses across training temperatures $\tau$.}
    \label{fig:defense_temp_acc_sece}
\end{figure}
\begin{figure}[t!]
    \centering
    \includegraphics[width=\linewidth]{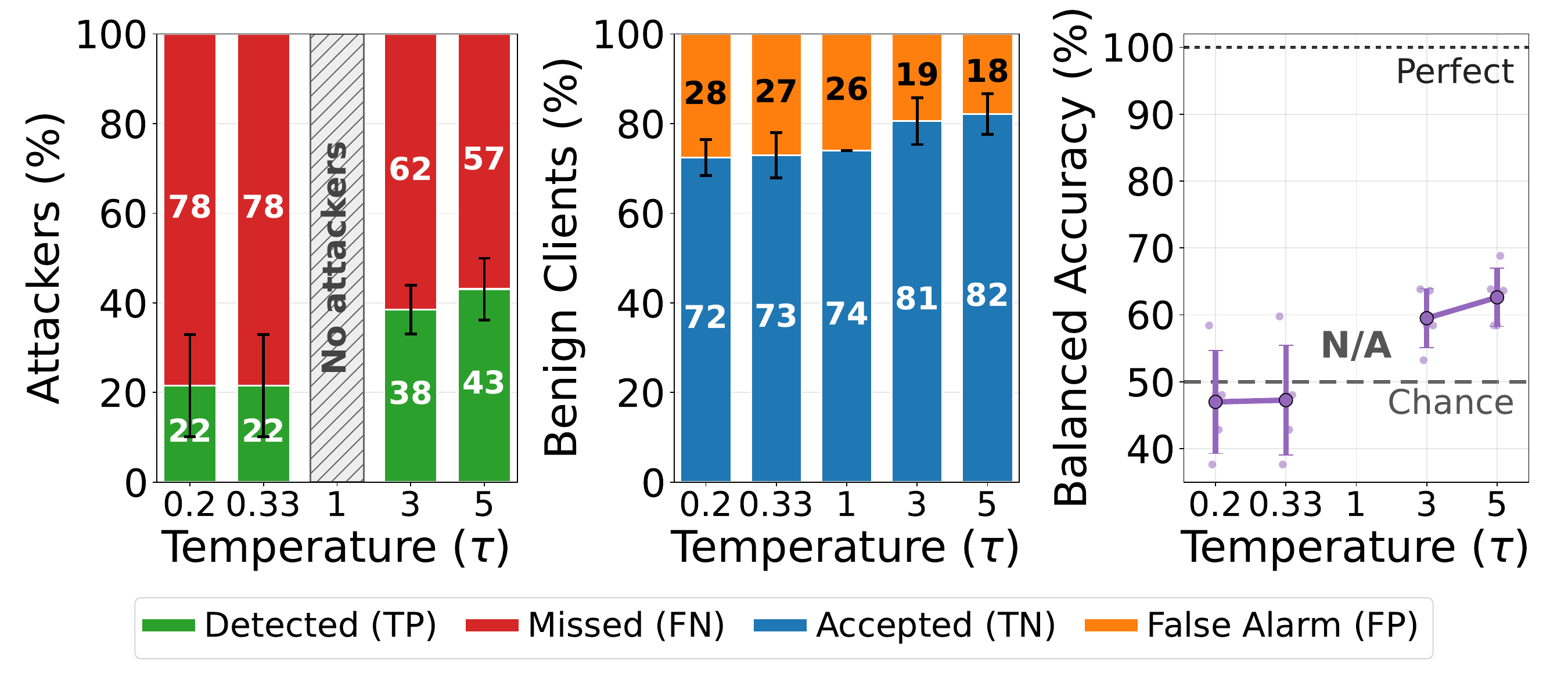}
    \caption{FLDetector performance against TSA across training temperatures. We report attacker detection, benign-client false alarms, and balanced accuracy.}
    \label{fig:detection-metric}
\end{figure}
\begin{figure*}[t!]
    \centering
    \includegraphics[width=.9\linewidth]{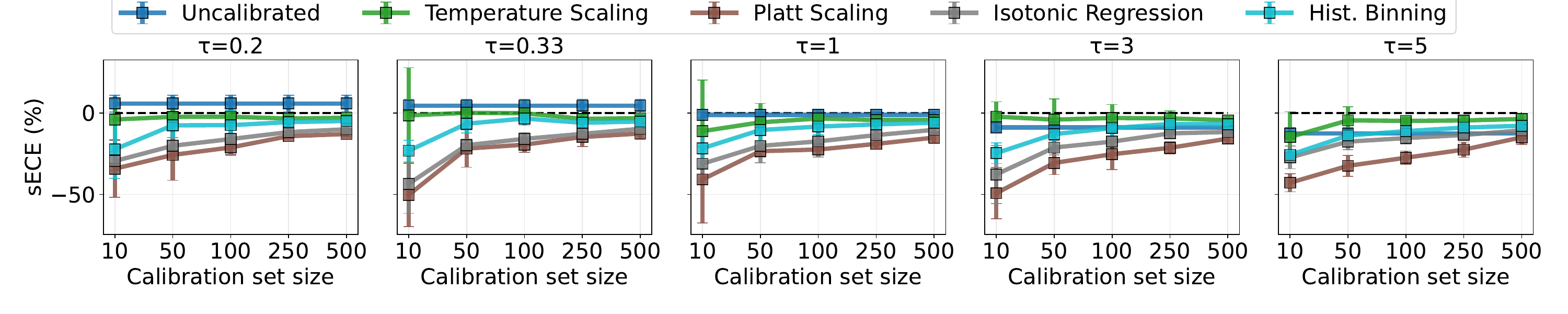}   
    \caption{Post-hoc calibration under temperature scaling attacks. We compare four calibration methods as a function of the number of calibration samples, reporting the resulting sECE after calibration.}
    \label{fig:posthoc-calibration}
\end{figure*}

\noindent{}\textbf{Sensitivity to Effective Step Size ($\beta$).}
Finally, Figure~\ref{fig:beta-acc-sece} studies how our key control knob (the effective step size $\beta=\eta/\tau$) affects both utility and calibration manipulation.
Note that $\beta$ coincides with the learning rate when $\tau{=}1$, so it naturally exhibits a task-dependent ``sweet spot'' for stable and accurate training.
Accordingly, varying $\beta$ induces some deviation in both accuracy and sECE across all conditions.
Nevertheless, the central takeaway remains unchanged: within a reasonable $\beta$ range, temperature scaling preserves accuracy close to the benign reference ($\tau{=}1$) while still steering sECE in the intended direction.
Specifically, $\tau<1$ consistently increases sECE (under-confidence) and $\tau>1$ decreases sECE (over-confidence), indicating that calibration control is robust to moderate changes in $\beta$ even though $\beta$ itself governs the overall optimization regime.

\subsection{Evaluation against Existing Defenses}
\label{subsec:defense}
We now evaluate the proposed temperature scaling attack under representative defenses used in practice.
We consider two broad axes: (i) \emph{server-side robust aggregation} that aims to suppress malicious client updates during training, and (ii) \emph{post-hoc calibration} that attempts to correct the confidence of the learned global model using a held-out validation set.

\noindent{}\textbf{Byzantine-robust defenses.}
Figure~\ref{fig:defense_temp_acc_sece} reports the global accuracy and sECE under six aggregation/defense schemes: FedAvg~\cite{mcmahan2017communication} (no defense), FLTrust~\cite{cao2020fltrust}, FoolsGold~\cite{fung2018mitigating}, MultiKrum~\cite{blanchard2017machine}, SureFED~\cite{heydaribeni2023surefed}, and FLDetector~\cite{zhang2022fldetector}.
The first three defenses primarily identify suspicious updates from their spatial relationships: FLTrust reweights updates according to their alignment with a trusted server update, FoolsGold penalizes clients with persistently similar update directions, and MultiKrum selects updates that are closest to the majority in parameter space.
We additionally include SureFED and FLDetector since they represent complementary defense principles beyond such direct update-space aggregation.
SureFED builds trust through uncertainty-aware evaluation using clients' locally trained reference models, while FLDetector detects malicious clients from temporal inconsistency between their observed and historically predicted updates.
Thus, together these methods cover update geometry, trusted-reference evaluation, uncertainty-aware inspection, and temporal update consistency.

Overall, none of these defenses consistently neutralizes TSA.
FoolsGold, MultiKrum, and SureFED substantially reduce model utility in our setting, yet sECE still changes systematically with $\tau$, indicating that calibration manipulation persists despite the accuracy degradation.
FLTrust preserves accuracy substantially better and partially attenuates over-confidence for $\tau>1$, but remains vulnerable to the under-confidence attack for $\tau<1$.
Similarly, FLDetector maintains accuracy close to FedAvg, but its sECE still shifts from positive values at $\tau<1$ to increasingly negative values at $\tau>1$.
This result is particularly notable because FLDetector explicitly targets malicious-client detection through update consistency rather than relying solely on a robust aggregation rule.
Taken together, the results show that TSA remains effective across defenses based on substantially different detection principles, suggesting that defenses designed primarily for accuracy-degrading model poisoning do not necessarily protect calibration integrity.

\noindent{}\textbf{Detectability under client-level monitoring.}
Beyond evaluating the resulting global model, we further examine whether TSA clients can be directly identified during federated training.
We use FLDetector~\cite{zhang2022fldetector} as a representative client-level detection method, as it identifies malicious participants by tracking temporal inconsistencies between observed client updates and their expected behavior.
Figure~\ref{fig:detection-metric} reports its client-level detection results across different attack temperatures.

Overall, detection becomes easier as the attack temperature moves farther from the benign setting, but TSA remains difficult to identify reliably.
For low-temperature attacks ($\tau{=}0.2,0.33$), approximately $78\%$ of attackers are missed, while $27$--$28\%$ of benign clients are falsely flagged, resulting in balanced accuracy of only about $47\%$.
For higher temperatures, detection improves but remains limited: the true-positive rate reaches only $38\%$ at $\tau{=}3$ and $43\%$ at $\tau{=}5$, leaving the majority of attackers undetected.
The corresponding balanced accuracies are $59.5\%$ and $62.6\%$, respectively.

Detection is also prone to false alarms and delay.
Even in the benign setting ($\tau{=}1$), where no attackers are present, approximately $26\%$ of clients are incorrectly flagged.
Moreover, the first actual attacker is typically detected only around rounds $38$--$44$.
These results indicate that update-consistency based client monitoring can partially expose stronger TSA configurations, but does not provide reliable detection across the attack range, particularly for low-temperature attacks.


\noindent{}\textbf{Post-hoc calibration.}
Finally, we evaluate whether post-hoc calibration, where the server or clients recalibrate predictive confidence using a held-out validation set, can mitigate the proposed attack. We consider four representative methods: temperature scaling~\cite{guo2017calibration}, Platt scaling~\cite{platt1999probabilistic}, isotonic regression~\cite{zadrozny2002transforming}, and histogram binning~\cite{zadrozny2001obtaining}. For each method, we vary the calibration set size from $\{10,50,100,250,500\}$ samples and measure the resulting sECE.

Figure~\ref{fig:posthoc-calibration} shows that increasing the number of calibration samples does not reliably neutralize the attack. Except for temperature scaling, the other post-hoc methods exhibit only limited improvement as more samples are available and, in several cases, even amplify signed miscalibration (larger-magnitude sECE). The most effective countermeasure is, somewhat paradoxically, temperature scaling itself; however, it still does not consistently recover clean calibration across attack temperatures. Moreover, in realistic federated deployments with non-IID data, acquiring representative validation data for calibration (and performing calibration repeatedly as the global model evolves) introduces non-trivial overhead and can be challenging in practice. These results suggest that post-hoc calibration alone is insufficient as a robust defense against training-time temperature manipulation.

\subsection{Case Study}
\label{subsec:casestudy}

The preceding experiments characterize TSA in controlled classification settings, focusing on confidence manipulation, predictive utility, optimization behavior, robustness, and existing defenses. We next ask whether such confidence shifts can affect downstream decisions in systems that directly consume model probabilities.

To examine this question, we consider three case studies: (i) mobile healthcare for remote monitoring, (ii) perception in robotics and autonomous driving, and (iii) text generation with language models. These studies are intended to illustrate how training-time confidence manipulation can propagate to application-level behavior through thresholding, safety gating, and probabilistic decoding.


\begin{figure}[t!]
    \centering
    \includegraphics[width=.95\linewidth]{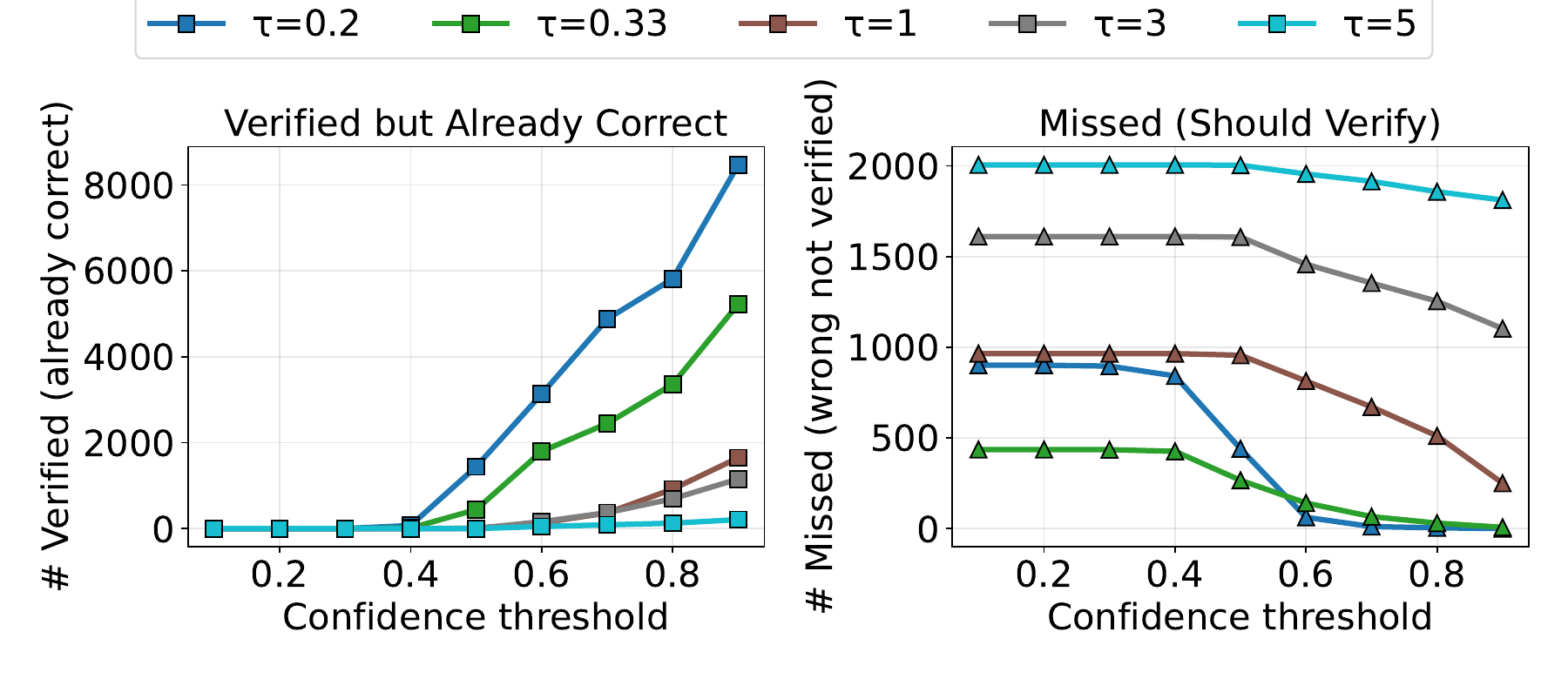}
    \caption{Impact of calibration distortion on threshold-based verification in a mobile healthcare triage workflow. Left: number of \emph{unnecessary} verifications (correct predictions that are still sent for verification). Right: number of \emph{missed} verifications (incorrect predictions that should have been verified but were not), as a function of the confidence threshold $\gamma$.}
    \label{fig:healthcare-threshold}
\end{figure}

\noindent{}\textbf{Case \#1: Mobile healthcare remote monitoring (risk-triggered triage).}
Mobile health applications increasingly rely on federated classifiers to detect clinically relevant events (e.g., arrhythmia episodes), where predictive confidence drives operational rules such as ``send to clinician if $p(y{=}\mathrm{abnormal}\mid x)>\gamma$ '' or ``verify low-confidence samples.'' Temperature scaling attacks are particularly damaging in such scenarios given that they can shift the probability scale without an obvious accuracy drop, directly corrupting threshold-based triage and resource allocation.

To quantify this effect, we train a 1D-CNN arrhythmia classifier with $10$ clients on the MIT-BIH arrhythmia database~\cite{moody2001impact} for $20$ FL rounds, and assume that samples with confidence below a threshold $\gamma$ are forwarded to a server/clinician for additional verification. Figure~\ref{fig:healthcare-threshold} plots two operational costs as $\gamma$ varies: (i) \emph{Verified but already correct}, i.e., unnecessary verification requests, and (ii) \emph{Missed (should verify)}, i.e., incorrect predictions that bypass verification. Under an under-confidence attack ($\tau<1$), the model assigns systematically lower confidences, which inflates the verification workload: at $\gamma{=}0.9$, unnecessary verifications increase from $1{,}647$ ($\tau{=}1$) to $8{,}465$ ($\tau{=}0.2$), a $+414.0\%$ rise.
Conversely, under an over-confidence attack ($\tau>1$), the model becomes overly certain, causing risky samples to skip verification: at $\gamma{=}0.9$, missed verifications increase from $250$ ($\tau{=}1$) to $1{,}812$ ($\tau{=}5$), a $+624.8\%$ rise.
These results suggest that calibration integrity is safety-critical in risk-triggered workflows, with the primary harm not being significant changes in accuracy, but a large shift in downstream decisions governed by confidence thresholds.

\begin{figure}[t!]
    \centering
    \includegraphics[width=.95\linewidth]{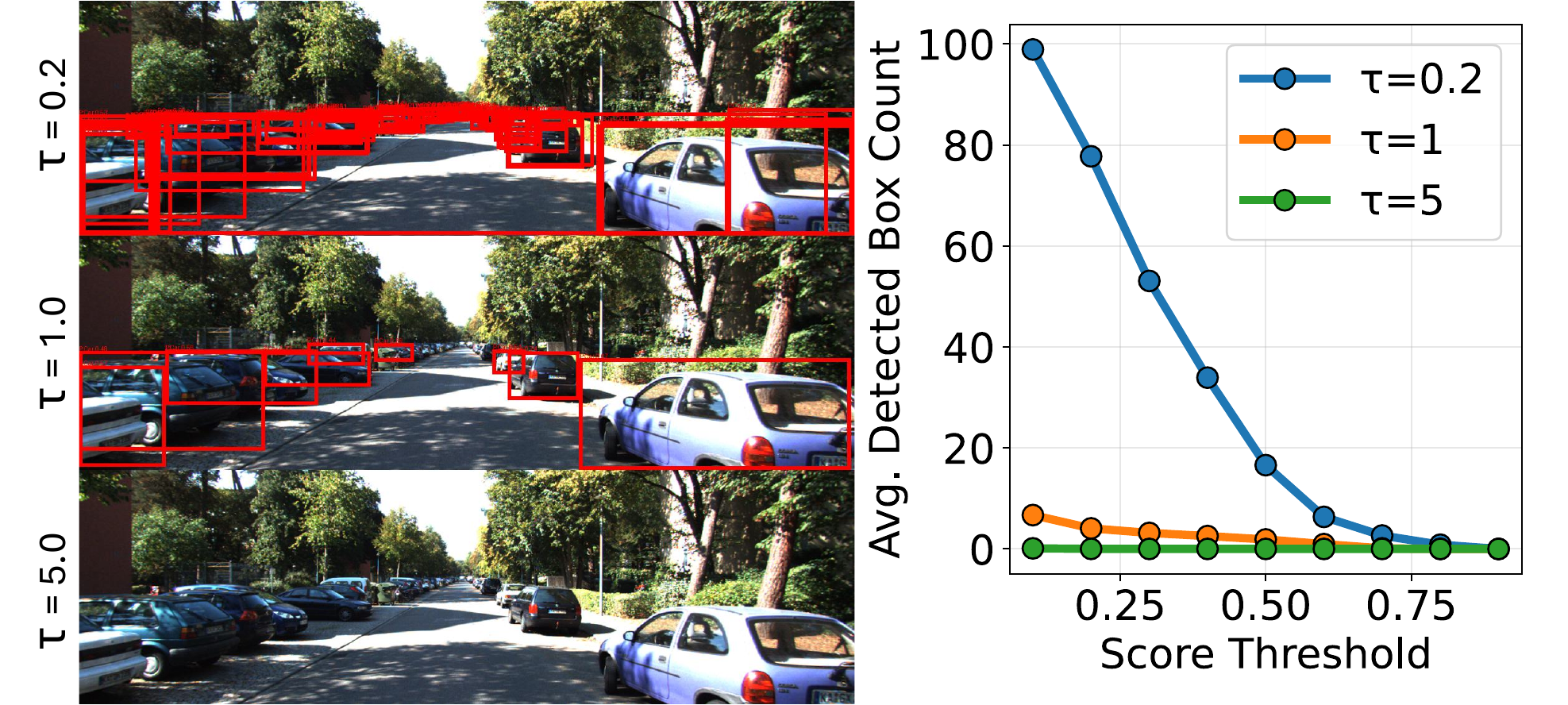}
    \caption{Selective perception under temperature-induced miscalibration in DETR. Left: detections on the same KITTI image at a fixed score threshold ($0.3$) for $\tau\in\{0.2,1,5\}$. Right: average number of retained boxes versus the score threshold. Low $\tau$ yields many false positives (too many boxes), while high $\tau$ suppresses detections (near-zero boxes).}
    \label{fig:detr-boxcount}
\end{figure}

\noindent{}\textbf{Case \#2: Robotics and autonomous driving (selective perception and safety gating).}
Robots and autonomous vehicles commonly use confidence scores from perception modules (e.g., detection/segmentation) to gate downstream behaviors, such as triggering conservative planning, requesting additional sensing, or initiating a fallback mode when confidence is low.
Temperature-driven miscalibration can systematically bias these gates as over-confidence may suppress fallbacks and pass incorrect outputs to planning, while under-confidence can cause excessive caution or unnecessary interventions. Since these policies react to calibrated confidence rather than accuracy alone, the impact can be disproportionate even when standard validation accuracy appears unchanged.

To emulate this scenario, we fine-tune a COCO2017-pretrained DETR model~\cite{carion2020end} on the KITTI 2D detection benchmark~\cite{Geiger2012CVPR} in a federated manner under three temperatures, $\tau\in\{0.2,1.0,5.0\}$. DETR produces $100$ object queries and, for each query, estimates a ``no-object'' probability; a query is kept as a detection only if its no-object score is below a threshold.
As a result, the temperature directly alters the effective selectivity of the detector.
With low temperature ($\tau{<}1$), the no-object probability becomes \emph{under-confident}; thus, many queries are incorrectly treated as objects, leading to a flood of detections.
With high temperature ($\tau{>}1$), the no-object score becomes \emph{over-confident}, leading to true objects being filtered.

Figure~\ref{fig:detr-boxcount} visualizes this failure mode. At the same score threshold ($0.3$), $\tau{=}0.2$ produces an excessive number of boxes that can overwhelm downstream tracking and planning, whereas $\tau{=}5$ can suppress detection altogether, effectively making the system ``blind.'' The right panel further confirms a systematic shift in the average number of retained boxes across thresholds. These results showcase how temperature scaling can translate into practical safety and performance risks by corrupting confidence-based gating in real perception pipelines.

\begin{figure}[t]
\centering
\begin{minipage}{0.99\linewidth}
\sloppy
\small
\noindent\fbox{%
\begin{minipage}{0.97\linewidth}
\sloppy
\textbf{Prompt (Input).}\ \texttt{Perhaps because Abraham Lincoln had not yet been}

\medskip
\textbf{$\tau=1.0$ (baseline).}\ 
\emph{Perhaps because Abraham Lincoln had not yet been inaugurated as President, Captain Totten received no instructions from his superiors.}

\medskip
\textbf{$\tau=0.5$ (attacked / under-confident).}\ 
\emph{Perhaps because Abraham Lincoln had not yet been inaugurated as President, not as President the arsenal of Captain Totten was constructed. On January 8, 1861, the state legislature decided a arsenal, but found no citizens to engage in such undertakings A grid @-@@@ person, not as this character. @-@ for @. Note there, then the which'Aotten received in the, construction.}
\end{minipage}}
\end{minipage}
\caption{Qualitative example of temperature-controlled generation behavior. Lower temperature ($\tau{=}0.5$) produces a degraded continuation with repetition, factual drift, and corrupted tokens, compared to the coherent baseline at $\tau{=}1.0$.}
\label{fig:temp_text_example}
\end{figure}

\textbf{Case \#3: Federated language models (confidence-based generation control).}
In federated language modeling, token probabilities are directly consumed by decoding and control logic (e.g., top-$p$ sampling~\cite{holtzman2019curious}, abstention, or retrieval triggers)~\cite{sani2025photon}. Thus, manipulating calibration during training can change generation behavior even with small next-token accuracy changes. In particular, lower temperature ($\tau{<}1$) training induces \emph{under-confidence}, yielding a \emph{flatter} predictive distribution at inference; conversely, higher temperatures push the model toward over-confidence and overly peaky posteriors.

We demonstrate this effect with a pretrained GPT-2~\cite{radford2019language} fine-tuned on WikiText-2~\cite{merity2016pointer} via FL. Using standard top-$p$ sampling at inference, the attacked model trained with $\tau{=}0.5$ produces markedly degraded continuations compared to the benign baseline ($\tau{=}1.0$), including repetition, factual drift, and corrupted tokens (Figure~\ref{fig:temp_text_example}). This highlights that training-time temperature manipulation can shift the probability scale that governs decoding, making confidence-based generation pipelines a practical target.

\section{Discussion}
\label{sec:discussion}

\noindent{}\textbf{Calibration Attacks in Federated Learning.}
We establish \emph{calibration integrity} as a distinct attack objective in federated learning.
While TSA provides one concrete realization, we hope it motivates broader research on calibration-oriented attacks and defenses.
As predictive confidence increasingly drives selective prediction, human review, and other downstream decisions, FL systems need calibration-aware protection beyond conventional accuracy- or loss-based monitoring.

\noindent{}\textbf{Practical Limitations.}
Our threat model assumes that attackers can control their local training pipeline, which is common in FL security studies but may require elevated capabilities such as rooting, jailbreaking, or bypassing integrity checks.
Future work should consider weaker adversaries that approximate similar calibration manipulation through data or label poisoning without directly modifying training.
Nevertheless, our results demonstrate the underlying vulnerability that calibration can be manipulated largely independently of accuracy.

\noindent{}\textbf{Future Work.}
Our systematic evaluation focuses primarily on classification, while the detection and language-model case studies provide initial evidence that the same confidence vulnerability can extend to other confidence-sensitive tasks. Future work should develop task-specific attack objectives and calibration measures for detection, segmentation, and sequence generation, and evaluate these settings across broader models and deployment conditions.
\section{Conclusion}
\label{sec:conclusion}

In this work, we demonstrated that \emph{calibration integrity} is a practical attack target in federated learning.
Our \emph{temperature scaling attack} manipulates predictive confidence during local training, steering the global model toward systematic overconfidence or underconfidence while largely preserving accuracy.
We analyzed temperature-scaled local SGD in the standard non-convex FL framework and showed that coupling the learning rate and temperature through $\beta=\eta/\tau$ controls the primary update scale, while leaving a bounded temperature-induced residual that captures the remaining deviation from benign optimization.
Across three benchmarks, the attack induces large, directional calibration shifts, yet remains difficult to flag using common signals such as accuracy trajectories, update-direction similarity, and representation/output similarity measures.
These results motivate calibration-aware monitoring and defenses in federated learning.

\bibliographystyle{IEEEtran}
\bibliography{reference}
\section*{Ethics Considerations}
This work introduces a new attack against calibration integrity in federated learning, which could potentially be misused to manipulate predictive confidence in deployed systems. However, our goal is to proactively identify and characterize this previously under-explored attack surface before it is exploited in practice. We provide detailed analyses of the attack mechanism, its limitations, and its detectability, and further evaluate representative defenses, including Byzantine-robust aggregation and post-hoc calibration. We believe that exposing these vulnerabilities and clarifying the conditions under which existing defenses succeed or fail provides greater defensive value than potential misuse, and can motivate the development of calibration-aware monitoring and protection mechanisms for future federated learning systems.
\appendix
\section*{Additional Theoretical Details}
\label{app:theory}

\subsection{Temperature-Dependent Smoothness}
\label{app:temp_smoothness}

We first characterize how temperature affects the curvature of softmax cross-entropy. Consider a single example $(x,y)$ with logits
$\mathbf{z}(\theta;x)\in\mathbb{R}^{C}$ and
\[
\mathbf{p}_{\tau}
\triangleq
\mathrm{softmax}(\mathbf{z}(\theta;x)/\tau).
\]
Writing
$\ell_{\tau}(\theta;x,y)
=
\tilde{\ell}_{\tau}(\mathbf{z}(\theta;x),y)$,
the Hessian with respect to the model parameters is
\begin{equation}
\nabla_{\theta}^{2}\ell_{\tau}
=
(\mathbf{J}_{\theta}\mathbf{z})^{\top}
\nabla_{\mathbf{z}}^{2}\tilde{\ell}_{\tau}
(\mathbf{J}_{\theta}\mathbf{z})
+
\sum_{i=1}^{C}
\frac{\partial\tilde{\ell}_{\tau}}{\partial z_i}
\nabla_{\theta}^{2}z_i.
\label{eq:hess_decomp_app}
\end{equation}

For softmax cross-entropy,
\begin{equation}
\nabla_{\mathbf{z}}^{2}\tilde{\ell}_{\tau}
=
\frac{1}{\tau^{2}}
\left(
\mathrm{diag}(\mathbf{p}_{\tau})
-
\mathbf{p}_{\tau}\mathbf{p}_{\tau}^{\top}
\right).
\end{equation}
The categorical covariance matrix has operator norm at most $1/2$. Hence, if
$\|\mathbf{J}_{\theta}\mathbf{z}\|\le G(x)$,
\begin{equation}
\left\|
(\mathbf{J}_{\theta}\mathbf{z})^{\top}
\nabla_{\mathbf{z}}^{2}\tilde{\ell}_{\tau}
(\mathbf{J}_{\theta}\mathbf{z})
\right\|
\le
\frac{G(x)^2}{2\tau^{2}}.
\label{eq:hess_termA_app}
\end{equation}

The second term follows from
\begin{equation}
\frac{\partial\tilde{\ell}_{\tau}}{\partial z_i}
=
\frac{1}{\tau}
\left(
p_{i,\tau}-\mathbbm{1}[i=y]
\right).
\end{equation}
Assuming
\begin{equation}
\sum_{i=1}^{C}
\|\nabla_{\theta}^{2}z_i\|
\le
H(x),
\end{equation}
we obtain
\begin{equation}
\left\|
\sum_{i=1}^{C}
\frac{\partial\tilde{\ell}_{\tau}}{\partial z_i}
\nabla_{\theta}^{2}z_i
\right\|
\le
\frac{H(x)}{\tau}.
\label{eq:hess_termB_app}
\end{equation}
Combining~\eqref{eq:hess_termA_app} and~\eqref{eq:hess_termB_app} gives
\begin{equation}
\|\nabla_{\theta}^{2}\ell_{\tau}\|
\le
\frac{G(x)^2}{2\tau^{2}}
+
\frac{H(x)}{\tau}.
\label{eq:hess_temp_bound_app}
\end{equation}

Assuming differentiation and expectation can be interchanged, taking expectation over the data gives
\begin{equation}
L_{\tau}
\le
\frac{\bar{G}^{2}}{2\tau^{2}}
+
\frac{\bar{H}}{\tau},
\end{equation}
where
$\bar{G}^{2}=\mathbb{E}[G(x)^2]$
and
$\bar{H}=\mathbb{E}[H(x)]$.
For logits that are affine in the optimized parameters, the second term vanishes and the bound reduces to $O(1/\tau^2)$.

Under the coupling $\eta_k=\beta\tau_k$,
\begin{equation}
\eta_k L_{\tau_k}
\le
\beta
\left(
\frac{\bar{G}^{2}}{2\tau_k}
+
\bar{H}
\right).
\label{eq:coupled_stability_app}
\end{equation}
Thus, for $\tau_k\ge\tau_{\min}>0$, a sufficiently small $\beta$ provides a uniform step-size stability condition over the considered temperature range. In particular, the bound becomes less favorable as $\tau_k$ approaches zero, consistent with the instability observed for very small temperatures. For piecewise-smooth networks, the same argument applies at points where the required derivatives exist. For the subsequent analysis, it is sufficient to choose $\beta$ such that
\begin{equation}
\beta E
\max\left\{
L,\,
\frac{\bar G^2}{2\tau_{\min}}+\bar H
\right\}
\le c_0,
\label{eq:stability_condition_app}
\end{equation}
for a sufficiently small constant $c_0>0$.

\subsection{Magnitude of the Temperature-Induced Residual}
\label{app:residual}

We next relate the residual in Eq.~\eqref{eq:gradient_decomposition} to the temperature itself. The Jacobian of the softmax mapping has operator norm at most $1/2$. Therefore,
\begin{equation}
\|\mathbf{p}_{\tau}-\mathbf{p}_{1}\|
\le
\frac{1}{2}
\left|
\frac{1}{\tau}-1
\right|
\|\mathbf{z}(\theta;x)\|.
\label{eq:softmax_temp_lipschitz}
\end{equation}
If additionally
$\|\mathbf{z}(\theta;x)\|\le Z(x)$
and
$\|\mathbf{J}_{\theta}\mathbf{z}(\theta;x)\|\le G(x)$,
then
\begin{equation}
\|\mathbf{r}_{\tau}(\theta;\xi)\|
\le
\frac{G(x)Z(x)}{2}
\left|
\frac{1}{\tau}-1
\right|.
\label{eq:residual_temp_bound_app}
\end{equation}

This bound is not required to select the attack temperature, but it provides intuition for the residual assumption in Eq.~\eqref{eq:residual_bound}. The residual is zero at $\tau=1$ and can grow as the training temperature moves away from the benign setting, particularly for small $\tau$. Thus, the $\eta$--$\tau$ coupling controls the explicit gradient scale but does not make arbitrarily extreme temperatures benign-like.

\subsection{Proof Sketch for Theorem~\ref{thm:convergence}}
\label{app:proof}

We now show how the temperature-induced residual modifies the standard non-convex local-SGD argument.

For a sample $\xi=(x,y)$, let
$\mathbf{g}_{k,\tau_k}(\theta;\xi)$
denote the stochastic gradient computed with temperature $\tau_k$, and let
$\mathbf{g}_{k}(\theta;\xi)$
denote the corresponding gradient at $\tau=1$.
From the softmax cross-entropy gradient,
\begin{equation}
\tau_k
\mathbf{g}_{k,\tau_k}(\theta;\xi)
=
\mathbf{g}_{k}(\theta;\xi)
+
\mathbf{r}_{k,\tau_k}(\theta;\xi),
\label{eq:gradient_decomp_app}
\end{equation}
where
\begin{equation}
\mathbf{r}_{k,\tau_k}(\theta;\xi)
=
(\mathbf{J}_{\theta}\mathbf{z})^{\top}
(\mathbf{p}_{\tau_k}-\mathbf{p}_{1}).
\end{equation}
Hence, using $\eta_k=\beta\tau_k$, each local step can be written exactly as
\begin{equation}
\theta_{k,t,e+1}
=
\theta_{k,t,e}
-
\beta
\left[
\mathbf{g}_{k}(\theta_{k,t,e};\xi_{k,t,e})
+
\mathbf{r}_{k,\tau_k}(\theta_{k,t,e};\xi_{k,t,e})
\right].
\label{eq:local_update_residual_app}
\end{equation}
Thus, coupled temperature-scaled SGD is equivalent to base local SGD with step size $\beta$ plus a temperature-dependent perturbation.

\paragraph{Client sampling and aggregation.}
For analysis, we define $\theta_{k,t,e}$ for every client as the counterfactual local trajectory it would produce from $\theta_t$ using its local stochastic gradients, regardless of whether client $k$ is selected in round $t$. Client sampling then selects among these local trajectories.

Let $\mathcal{K}_t$ denote the clients selected in round $t$, and let
$\mathcal{A}_t[\{\mathbf{v}_k\}]$
denote the weighted aggregate of client vectors.
We assume that, for any fixed collection $\{\mathbf{v}_k\}$,
\begin{equation}
\mathbb{E}_{\mathcal{K}_t}
\left[
\mathcal{A}_t[\{\mathbf{v}_k\}]
\right]
=
\sum_{k=1}^{K}p_k\mathbf{v}_k,
\label{eq:agg_unbiased_app}
\end{equation}
and that the sampling variance satisfies
\begin{equation}
\mathbb{E}_{\mathcal{K}_t}
\left[
\left\|
\mathcal{A}_t[\{\mathbf{v}_k\}]
-
\sum_{k=1}^{K}p_k\mathbf{v}_k
\right\|^2
\right]
\le
\frac{c_s}{M}
\sum_{k=1}^{K}
p_k
\left\|
\mathbf{v}_k-\bar{\mathbf{v}}
\right\|^2,
\label{eq:agg_variance_app}
\end{equation}
for a constant $c_s$ and
$\bar{\mathbf{v}}=\sum_kp_k\mathbf{v}_k$.
This condition captures the usual $O(1/M)$ variance reduction under unbiased partial participation.

The global update is
\begin{equation}
\Delta_t
\triangleq
\theta_{t+1}-\theta_t
=
-\beta
\sum_{e=0}^{E-1}
\mathcal{A}_t
\left[
\mathbf{g}_{k}(\theta_{k,t,e})
+
\mathbf{r}_{k,\tau_k}(\theta_{k,t,e})
\right].
\label{eq:global_update_app}
\end{equation}

Since $F$ is $L$-smooth,
\begin{equation}
F(\theta_{t+1})
\le
F(\theta_t)
+
\langle\nabla F(\theta_t),\Delta_t\rangle
+
\frac{L}{2}\|\Delta_t\|^2.
\label{eq:descent_lemma_app}
\end{equation}

\paragraph{Mean update and local drift.}
Define
\begin{equation}
\mathbf{d}_{t,e}
\triangleq
\sum_{k=1}^{K}
p_k
\left[
\nabla f_k(\theta_{k,t,e})
-
\nabla f_k(\theta_t)
\right]
\end{equation}
and
\begin{equation}
\mathbf{b}_{t,e}
\triangleq
\sum_{k=1}^{K}
p_k
\mathbb{E}_{\xi}
\left[
\mathbf{r}_{k,\tau_k}(\theta_{k,t,e};\xi)
\right].
\end{equation}
Using the unbiasedness of the base stochastic gradients and of client sampling,
\begin{equation}
\mathbb{E}[\Delta_t\mid\theta_t]
=
-\beta
\sum_{e=0}^{E-1}
\left(
\nabla F(\theta_t)
+
\mathbf{d}_{t,e}
+
\mathbf{b}_{t,e}
\right).
\label{eq:update_mean_app}
\end{equation}

By Jensen's inequality and the client-wise residual bound in Eq.~\eqref{eq:residual_bound},
\begin{equation}
\mathbb{E}\|\mathbf{b}_{t,e}\|^2
\le
\sum_{k=1}^{K}p_k\rho_k^2
\le
\rho^2.
\label{eq:residual_mean_bound_app}
\end{equation}
Young's inequality then gives
\begin{equation}
|\langle\nabla F(\theta_t),\mathbf{b}_{t,e}\rangle|
\le
\epsilon\|\nabla F(\theta_t)\|^2
+
\frac{1}{4\epsilon}
\|\mathbf{b}_{t,e}\|^2,
\label{eq:young_residual_app}
\end{equation}
so the residual contributes $O(\beta E\rho^2)$ to the one-round recursion.

For the ordinary local drift, smoothness gives
\begin{equation}
\|\mathbf{d}_{t,e}\|^2
\le
L^2
\sum_{k=1}^{K}
p_k
\|\theta_{k,t,e}-\theta_t\|^2.
\label{eq:local_drift_smooth_app}
\end{equation}
Standard local-SGD recursion applied to~\eqref{eq:local_update_residual_app} yields, for a sufficiently small stable $\beta$,
\begin{equation}
\begin{aligned}
\frac{1}{E}
\sum_{e=0}^{E-1}
\sum_{k=1}^{K}
p_k
\mathbb{E}
\|\theta_{k,t,e}-\theta_t\|^2
=\\
O\left(
\beta^2E^2
\left[
\|\nabla F(\theta_t)\|^2
+
\Gamma^2
+
\sigma^2
+
\rho^2
\right]
\right).
\label{eq:local_distance_bound_app}
\end{aligned}
\end{equation}
The term involving $\|\nabla F(\theta_t)\|^2$ is absorbed into the main descent term under the same step-size condition, and the $\rho^2$ contribution can be absorbed into the final $O(\rho^2)$ term.

\paragraph{Variance terms.}
There are two standard sources of sampling variance. Mini-batch noise contributes
\begin{equation}
O\left(
\frac{\beta^2LE\sigma^2}{M}
\right)
\end{equation}
to the one-round recursion. In addition, because the same sampled clients participate across the $E$ local steps, partial client participation contributes
\begin{equation}
O\left(
\frac{\beta^2LE^2\Gamma^2}{M}
\right).
\end{equation}
The latter term is the sampling variance induced by heterogeneous client objectives and vanishes under full participation.

Combining these terms with the local-drift and residual bounds gives
\begin{equation}
\begin{aligned}
\mathbb{E}[F(\theta_{t+1})]
\le\;&
\mathbb{E}[F(\theta_t)]
-
\frac{\beta E}{2}
\mathbb{E}
\left[
\|\nabla F(\theta_t)\|^2
\right]
\\
&+
O\left(
\frac{\beta^2LE\sigma^2}{M}
\right)
+
O\left(
\frac{\beta^2LE^2\Gamma^2}{M}
\right)
\\
&+
O\left(
\beta^3L^2E^3(\Gamma^2+\sigma^2)
\right)
+
O(\beta E\rho^2).
\end{aligned}
\label{eq:one_round_recursion_app}
\end{equation}

Here, terms of order
$\beta^3L^2E^3\rho^2$
arising through the local trajectory are absorbed into
$O(\beta E\rho^2)$ under the stable-step condition.

Summing~\eqref{eq:one_round_recursion_app} over
$t=0,\ldots,T-1$,
using
$F(\theta_T)\ge F^\star$,
and dividing by $\beta ET/2$ gives
\begin{equation}
\begin{aligned}
\frac{1}{T}
\sum_{t=0}^{T-1}
\mathbb{E}
\left[
\|\nabla F(\theta_t)\|^2
\right]
\le\;&
\frac{2(F(\theta_0)-F^\star)}
{\beta ET}
\\
&+
O\left(
\frac{\beta L\sigma^2}{M}
\right)
+
O\left(
\frac{\beta LE\Gamma^2}{M}
\right)
\\
&+
O\left(
\beta^2L^2E^2(\Gamma^2+\sigma^2)
\right)
+
O(\rho^2).
\end{aligned}
\label{eq:convergence_bound_app}
\end{equation}

Equivalently, the two partial-participation variance terms can be written compactly as
\begin{equation}
O\left(
\frac{\beta L(\sigma^2+E\Gamma^2)}{M}
\right).
\end{equation}
This yields the convergence bound stated in Theorem~\ref{thm:convergence}.

When $\tau_k=1$ for every client, the residual vanishes and the standard benign local-SGD form is recovered. For $\tau_k\neq1$, the coupling removes the explicit $1/\tau_k$ gradient-scale factor, while $\mathbf{r}_{k,\tau_k}$ captures the remaining temperature-dependent change in the update. Thus, the additional convergence penalty is governed by the magnitude of this residual, which itself depends on how far the temperature moves the predictive distribution away from its benign counterpart.

\subsection{Extension to Momentum SGD and Adam-Type Optimizers}
\label{app:adaptive_optimizers}

The main convergence analysis considers local SGD. We briefly discuss how the temperature effect changes under momentum SGD and Adam-type optimizers.

\paragraph{Momentum SGD.}
For a fixed local temperature $\tau$, recall the exact gradient decomposition
\begin{equation}
\mathbf{g}_t^{(\tau)}
=
\frac{1}{\tau}
\left(
\mathbf{g}_t^{(1)}
+
\mathbf{r}_t^{(\tau)}
\right)
\triangleq
\frac{1}{\tau}\mathbf{q}_t^{(\tau)},
\label{eq:momentum_temp_grad}
\end{equation}
where all quantities are evaluated along the same temperature-scaled local trajectory.
Consider the momentum recursion
\begin{equation}
\mathbf{u}_t^{(\tau)}
=
\mu\mathbf{u}_{t-1}^{(\tau)}
+
\mathbf{g}_t^{(\tau)},
\qquad
\theta_{t+1}
=
\theta_t-\eta_\tau\mathbf{u}_t^{(\tau)}.
\end{equation}
With a zero-initialized momentum buffer and a fixed $\tau$,
\begin{equation}
\mathbf{u}_t^{(\tau)}
=
\frac{1}{\tau}
\mathbf{u}_t[\mathbf{q}^{(\tau)}].
\end{equation}
Hence, under $\eta_\tau=\beta\tau$,
\begin{equation}
\eta_\tau\mathbf{u}_t^{(\tau)}
=
\beta\mathbf{u}_t[\mathbf{q}^{(\tau)}].
\end{equation}
Thus, the explicit $1/\tau$ gradient scale is canceled through the momentum recursion in the same manner as for SGD. Temperature dependence remains through
$\mathbf{q}_t^{(\tau)}
=
\mathbf{g}_t^{(1)}+\mathbf{r}_t^{(\tau)}$.

\paragraph{Adam-type optimization.}
Adam behaves differently because it normalizes each coordinate using running first- and second-moment estimates. Let
\begin{align}
\mathbf{m}_t^{(\tau)}
&=
\beta_1\mathbf{m}_{t-1}^{(\tau)}
+
(1-\beta_1)\mathbf{g}_t^{(\tau)},\\
\mathbf{v}_t^{(\tau)}
&=
\beta_2\mathbf{v}_{t-1}^{(\tau)}
+
(1-\beta_2)
\mathbf{g}_t^{(\tau)}\odot\mathbf{g}_t^{(\tau)},
\end{align}
with the usual bias-corrected moments
$\hat{\mathbf m}_t^{(\tau)}$ and
$\hat{\mathbf v}_t^{(\tau)}$.
For a fixed $\tau>0$ and zero-initialized moments,
Eq.~\eqref{eq:momentum_temp_grad} gives
\begin{equation}
\hat{\mathbf m}_t^{(\tau)}
=
\frac{1}{\tau}
\hat{\mathbf m}_t[\mathbf q^{(\tau)}],
\qquad
\hat{\mathbf v}_t^{(\tau)}
=
\frac{1}{\tau^2}
\hat{\mathbf v}_t[\mathbf q^{(\tau)}].
\end{equation}
Therefore, the normalized Adam direction satisfies
\begin{equation}
\mathbf d_t^{(\tau)}
\triangleq
\frac{\hat{\mathbf m}_t^{(\tau)}}
{\sqrt{\hat{\mathbf v}_t^{(\tau)}}+\epsilon}
=
\frac{\hat{\mathbf m}_t[\mathbf q^{(\tau)}]}
{\sqrt{\hat{\mathbf v}_t[\mathbf q^{(\tau)}]}+\tau\epsilon}.
\label{eq:adam_temp_direction}
\end{equation}

Equation~\eqref{eq:adam_temp_direction} shows an important distinction from SGD. When $\epsilon=0$, Adam itself exactly removes a uniform $1/\tau$ rescaling of the gradient. For the usual small positive $\epsilon$, this scaling is still largely normalized. Temperature nevertheless changes the normalized descent direction through
$\mathbf q_t^{(\tau)}$, because
$\mathbf{r}_t^{(\tau)}$ changes the relative class-wise and coordinate-wise gradient structure. Consequently, the SGD interpretation of $\eta/\tau$ as an exact effective-step-size invariance does not directly extend to Adam.

We can nevertheless obtain a standard descent guarantee for the resulting Adam direction.

\begin{proposition}[Descent under temperature-scaled Adam]
\label{prop:adam_convergence}
Consider
\begin{equation}
\theta_{t+1}
=
\theta_t-\eta_\tau\mathbf d_t^{(\tau)},
\end{equation}
where $\mathbf d_t^{(\tau)}$ is defined in
Eq.~\eqref{eq:adam_temp_direction}.
Suppose $F$ is $L$-smooth and bounded below by $F^\star$, and assume that, for constants
$c_A>0$, $G_A>0$, and $\delta_A^2\ge0$,
\begin{align}
\mathbb{E}\!\left[
\|\mathbf d_t^{(\tau)}\|^2
\mid \mathcal{F}_t
\right]
&\le
G_A^2,
\label{eq:adam_direction_bound}\\
\mathbb{E}\!\left[
\left\langle
\nabla F(\theta_t),
\mathbf d_t^{(\tau)}
\right\rangle
\mid \mathcal{F}_t
\right]
&\ge
c_A\|\nabla F(\theta_t)\|^2-\delta_A^2,
\label{eq:adam_alignment}
\end{align}
where $\mathcal{F}_t$ denotes the optimization history up to step $t$.
Then, for a constant learning rate $\eta_\tau>0$,
\begin{equation}
\frac{1}{T}
\sum_{t=0}^{T-1}
\mathbb{E}
\left[
\|\nabla F(\theta_t)\|^2
\right]
\le
\frac{F(\theta_0)-F^\star}
{c_A\eta_\tau T}
+
\frac{\delta_A^2}{c_A}
+
\frac{L\eta_\tau G_A^2}{2c_A}.
\label{eq:adam_convergence}
\end{equation}
\end{proposition}

\paragraph{Proof sketch.}
By $L$-smoothness of the base objective,
\begin{equation}
F(\theta_{t+1})
\le
F(\theta_t)
-
\eta_\tau
\left\langle
\nabla F(\theta_t),
\mathbf d_t^{(\tau)}
\right\rangle
+
\frac{L\eta_\tau^2}{2}
\|\mathbf d_t^{(\tau)}\|^2.
\end{equation}
Taking conditional expectation and applying
Eqs.~\eqref{eq:adam_direction_bound}--\eqref{eq:adam_alignment} gives
\begin{equation}
\begin{aligned}
\mathbb{E}[F(\theta_{t+1})\mid\mathcal{F}_t]
\le\;&
F(\theta_t)
-
c_A\eta_\tau
\|\nabla F(\theta_t)\|^2\\
&+
\eta_\tau\delta_A^2
+
\frac{L\eta_\tau^2G_A^2}{2}.
\end{aligned}
\end{equation}
Summing over $t$, using $F(\theta_T)\ge F^\star$, and dividing by
$c_A\eta_\tau T$ yields
Eq.~\eqref{eq:adam_convergence}.

Proposition~\ref{prop:adam_convergence} is distinct from the SGD result: it does not claim an exact $\eta/\tau$ collapse for Adam. Instead, Adam internally normalizes the explicit temperature-dependent gradient scale, while temperature continues to affect the normalized descent direction through
$\mathbf{r}_t^{(\tau)}$.
If this direction remains sufficiently aligned with the base objective
(i.e., $c_A>0$ and $\delta_A$ remains bounded), temperature-scaled Adam retains a standard descent guarantee while still allowing model confidence to change.

If the same coupling $\eta_\tau=\beta\tau$ is used with Adam, Eq.~\eqref{eq:adam_convergence} becomes
\begin{equation}
\frac{1}{T}
\sum_{t=0}^{T-1}
\mathbb{E}
\left[
\|\nabla F(\theta_t)\|^2
\right]
\le
\frac{F(\theta_0)-F^\star}
{c_A\beta\tau T}
+
\frac{\delta_A^2}{c_A}
+
\frac{L\beta\tau G_A^2}{2c_A}.
\end{equation}
Thus, unlike SGD, fixing $\eta_\tau/\tau=\beta$ does not make the Adam update invariant to $\tau$; it instead requires $\beta\tau$ to remain within a stable learning-rate range. This optimizer-specific distinction is consistent with our use of the same temperature manipulation under both SGD-based and Adam-based local training.

For AdamW, the decoupled weight-decay component introduces an additional learning-rate-dependent update. The proposition above characterizes the adaptive gradient component; the weight-decay contribution can be treated separately and does not change the temperature-scaling identity in Eq.~\eqref{eq:adam_temp_direction}.

\section*{Experimental Details}
\label{app:exp_details}
\begin{table}[t]
\centering
\small
\setlength{\tabcolsep}{6pt}
\begin{adjustbox}{width=.9\linewidth, center}
\begin{tabular}{lccc}
\toprule
\textbf{Configuration} & \textbf{MNIST} & \textbf{CIFAR10} & \textbf{CIFAR100} \\
\midrule
\# Clients ($K$)          & \multicolumn{3}{c}{50} \\
Clients / round ($M$)     & \multicolumn{3}{c}{5 (10\%)} \\
Attacker ratio            & \multicolumn{3}{c}{25\% (13/50)} \\
Dirichlet $\alpha$        & \multicolumn{3}{c}{1.0} \\
Optimizer                 & \multicolumn{3}{c}{SGD} \\
Model architecture        & MLP  & CNN  & ResNet18 \\
Effective step ($\beta$)  & 0.01 & 0.05 & 0.05 \\
Batch size                & \multicolumn{3}{c}{64} \\
Total rounds              & \multicolumn{3}{c}{100} \\
Local steps ($E$)         & \multicolumn{3}{c}{5} \\
Random seeds              & \multicolumn{3}{c}{\{0,1,2,3,4\}} \\
\bottomrule
\end{tabular}
\end{adjustbox}
\caption{Federated learning and attack configuration across dataset--model pairs (default settings unless stated otherwise).}
\label{tab:exp_config}
\end{table}
\subsection{Federated learning setup}
\label{app:exp_fl_setup}
This appendix provides implementation and protocol details to facilitate reproducibility.
Unless otherwise specified, we use synchronous cross-device FL with FedAvg as summarized in Table~\ref{tab:exp_config}.
We instantiate $K=50$ total clients and sample $M=5$ clients (10\%) uniformly at random per round.
Each selected client performs $E=5$ steps of local SGD on its private data and sends model updates back to the server, which aggregates using data-size weights $p_k$.
\begin{table}[]
\begin{adjustbox}{width=.9\linewidth, center}
\begin{tabular}{lccc}
\toprule
\textbf{Configuration}   & \textbf{Case \#1}                                                & \textbf{Case \#2}                                                                   & \textbf{Case \#3}                                                                   \\
\midrule
Dataset                  & \multicolumn{1}{c}{MITBIH}                                       & \multicolumn{1}{l}{KITTI-2D}                                                        & \multicolumn{1}{c}{WikiText2}                                                       \\
Model architecture       & 1D CNN                                                           & DETR                                                                                & GPT-2                                                                               \\
Pretraining              & \multicolumn{1}{c}{\textcolor{red}{$\times$}} & \multicolumn{1}{c}{\textcolor{green}{\checkmark}} & \multicolumn{1}{c}{\textcolor{green}{\checkmark}} \\
\# Clients ($K$)         & 20                                                               & 10                                                                                  & 10                                                                                  \\
Clients / round ($M$)    & \multicolumn{3}{c}{5}                                                                                                                                                                                                                 \\
Attacker ratio           & \multicolumn{3}{c}{25\%}  \\
Optimizer                & SGD & \multicolumn{2}{c}{AdamW}                                                                                                                                                                                                                     \\
Effective step ($\beta$) & 0.01 & 1e-4 & 2e-3   \\
Batch size               & \multicolumn{3}{c}{64}                                                                                                                                                                                                                       \\
Total rounds             & \multicolumn{3}{c}{20}                                                                                                                                                                                                                      \\
Local steps ($E$)        & \multicolumn{3}{c}{5}                                                                                                                                                                                                                        \\
\bottomrule
\end{tabular}
\end{adjustbox}
\caption{Configuration used for case studies in Section~\ref{subsec:casestudy}.}
\label{tab:casestudy_config}
\end{table}

\begin{figure}[t!]
    \centering
    \includegraphics[width=.9\linewidth]{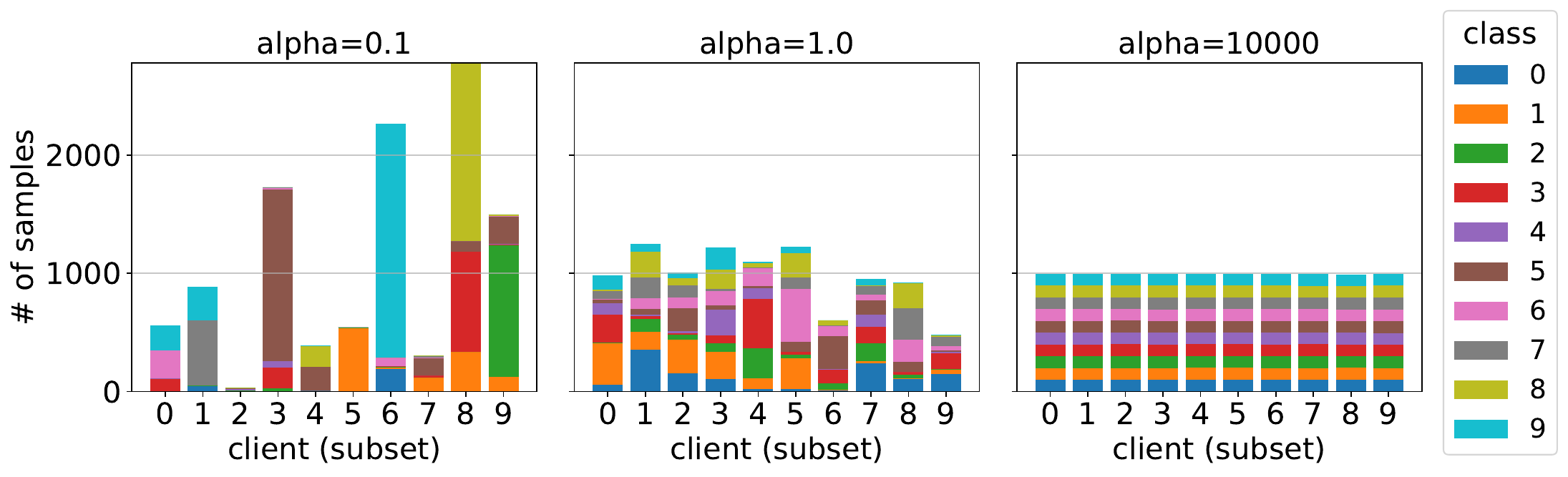}
    \caption{Example client label distributions under Dirichlet partitioning for CIFAR-10 with different concentration parameters $\alpha$ (smaller $\alpha$ yields more heterogeneous splits).}
    \label{fig:noniid-viz}
\end{figure}
\paragraph{Non-IID partitioning.}
To model statistical heterogeneity, we partition each dataset across clients using a Dirichlet split~\cite{hsu2019measuring}.
Specifically, for each class $c$, we draw a client-wise proportion vector $\boldsymbol{\pi}_c \sim \mathrm{Dirichlet}(\alpha)$ and allocate samples of class $c$ to clients according to $\boldsymbol{\pi}_c$.
The concentration parameter $\alpha$ controls the non-IID severity: smaller $\alpha$ yields more skewed, label-dominant clients, while larger $\alpha$ approaches an approximately IID split.
Figure~\ref{fig:noniid-viz} visualizes representative label distributions under different $\alpha$. As shown, the data distribution emulates both the quantity and label heterogeneity.
Unless otherwise specified, we use $\alpha=1.0$, and we sweep $\alpha$ in our sensitivity experiments (Fig.~\ref{fig:alpha-acc-sece}).

In addition, for the case studies in Section~\ref{subsec:casestudy}, we use different experiment configuration as shown in Table~\ref{tab:casestudy_config}.

\subsection{Centered Kernel Alignment}
\label{app:cka}
Centered Kernel Alignment (CKA) is a similarity measure between two sets of representations that compares their \emph{pairwise relational structure} rather than individual coordinates.
Given two feature matrices $X\in\mathbb{R}^{n\times d_x}$ and $Y\in\mathbb{R}^{n\times d_y}$ extracted from the same $n$ inputs, CKA measures how similarly $X$ and $Y$ represent the input set.

Direct cosine similarity between high-dimensional features is often brittle because it depends on the particular basis of the representation space.
In contrast, CKA is invariant to orthogonal transformations of the features (e.g., rotations) and is robust to feature reparameterizations that preserve representational structure.
This makes CKA well-suited for comparing penultimate-layer features across independently trained models, where equivalent representations may differ by a rotation or other benign reparameterizations.

In this work, we use \emph{linear} CKA, which is efficient and widely used for neural representation comparison.
Let $\tilde{X}$ and $\tilde{Y}$ denote column-centered features (i.e., each feature dimension has zero mean over the $n$ samples).
Linear CKA is defined as
\begin{equation}
\mathrm{CKA}(X,Y)
\triangleq
\frac{\|\tilde{X}^{\top}\tilde{Y}\|_{F}^{2}}
     {\|\tilde{X}^{\top}\tilde{X}\|_{F}\;\|\tilde{Y}^{\top}\tilde{Y}\|_{F}}.
\label{eq:linear_cka}
\end{equation}
Equivalently, linear CKA can be interpreted as the cosine similarity between the (vectorized) centered Gram matrices $\tilde{X}\tilde{X}^\top$ and $\tilde{Y}\tilde{Y}^\top$.
In our representation-space analysis, we compute CKA on penultimate-layer activations using a shared evaluation set, so that differences reflect representation drift rather than differences in input distribution.

\subsection{Defenses and post-hoc calibration}
\label{app:exp_defenses}
We evaluate representative robust aggregation rules (FedAvg, FLTrust, FoolsGold, MultiKrum) in Sec.~\ref{subsec:defense}.
When a defense introduces additional hyperparameters, we follow the authors' recommended settings and keep them fixed across temperatures so that the only varying factor is $\tau$.

\noindent{}\textbf{$\bullet$ Server-side defenses.}
\textbf{FedAvg}~\cite{mcmahan2017communication} is the standard baseline that aggregates client updates by data-size weights and performs no explicit adversary filtering.
\textbf{FLTrust}~\cite{cao2020fltrust} assigns \emph{trust scores} to client updates using a small server-held trusted dataset.
Concretely, the server computes a reference update on the trusted data and reweights each client update by its cosine similarity to the reference direction, together with norm normalization, suppressing updates that are poorly aligned with the trusted gradient.
\textbf{FoolsGold}~\cite{fung2018mitigating} targets sybil-style poisoning by downweighting clients whose updates are \emph{too similar} to others over time.
It maintains per-client similarity statistics and reduces the aggregation weight of clients that repeatedly submit highly correlated update directions, under the intuition that colluding attackers tend to produce clustered gradients.
\textbf{MultiKrum}~\cite{blanchard2017machine} is a distance-based Byzantine-robust rule.
For each client update, it computes a score based on distances to its closest neighboring updates and selects the updates with the smallest scores, corresponding to those most consistent with the majority.

\textbf{SureFED}~\cite{heydaribeni2023surefed} builds trust through uncertainty-aware inward and outward inspection using clients' locally trained reference models.
It evaluates whether received model information is consistent with each client's local observations and suppresses contributions considered unreliable.
\textbf{FLDetector}~\cite{zhang2022fldetector} detects malicious clients by exploiting temporal consistency in their submitted updates.
It predicts each client's expected update from its historical behavior and identifies clients whose observed updates deviate substantially from these predictions.

\begin{figure}[t!]
    \centering
    \includegraphics[width=.9\linewidth]{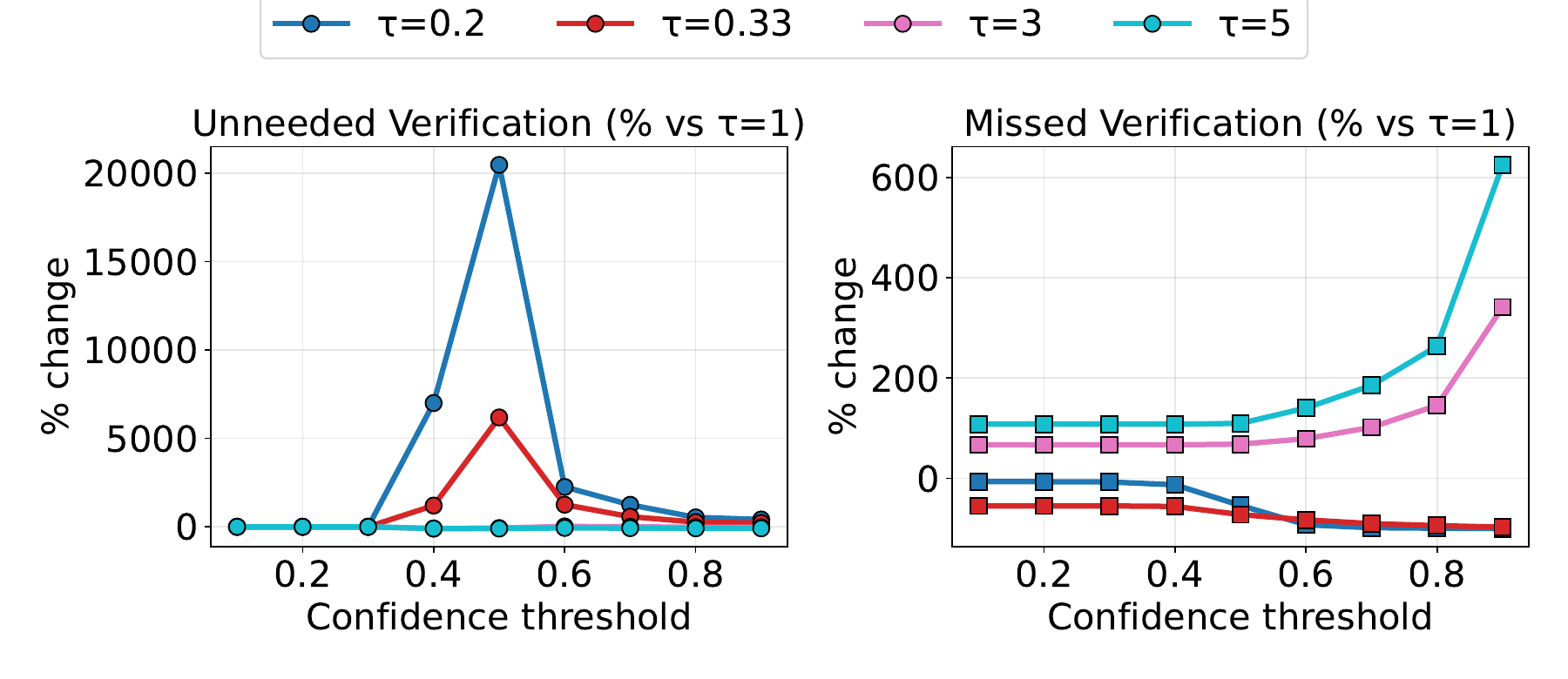}
    \caption{Impact of calibration distortion on threshold-based verification in mobile healthcare triage in percentage change (\%). Left: \emph{unnecessary} verifications. Right: \emph{missed} verifications, versus threshold $\gamma$.}
    \label{fig:healthcare-threshold-extend}
\end{figure}
\begin{figure}[t!]
    \centering
    \includegraphics[width=.9\linewidth]{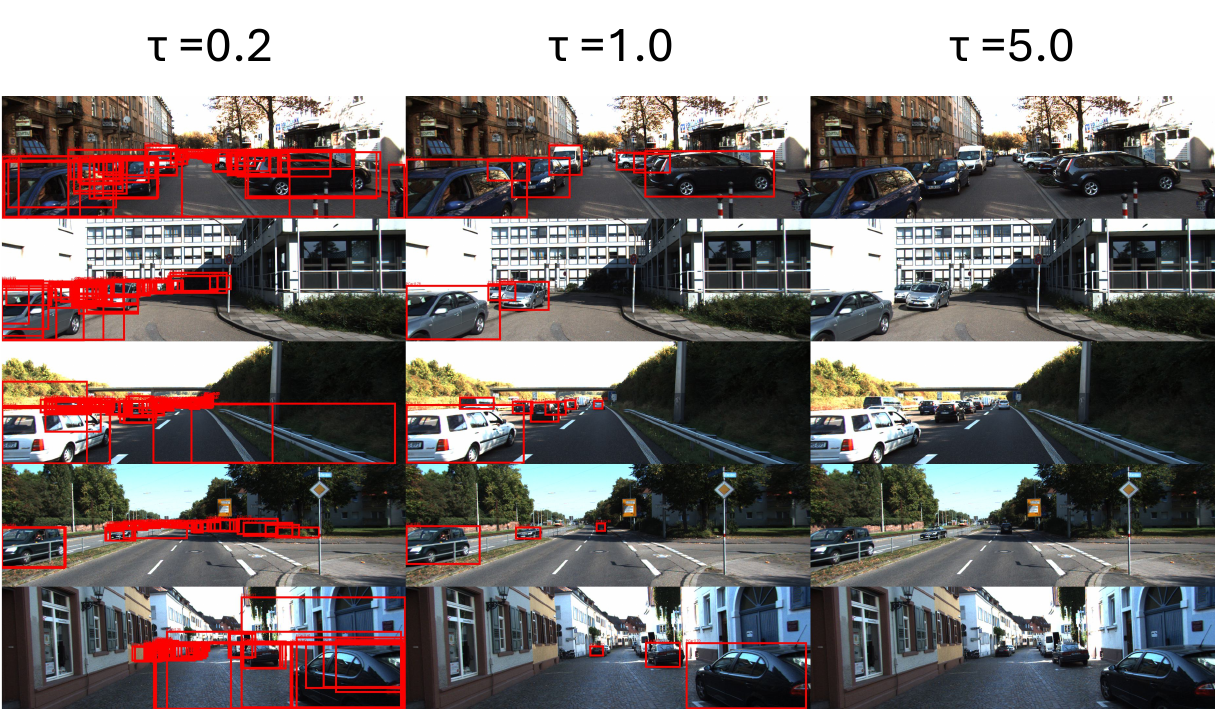}
    \caption{Extended results for selective perception under temperature-induced miscalibration in DETR.}
    \label{fig:detr-appendix}
\end{figure}
\begin{figure*}[t!]
    \centering

    \begin{subfigure}[t]{\linewidth}
        \centering
        \includegraphics[width=0.85\linewidth]{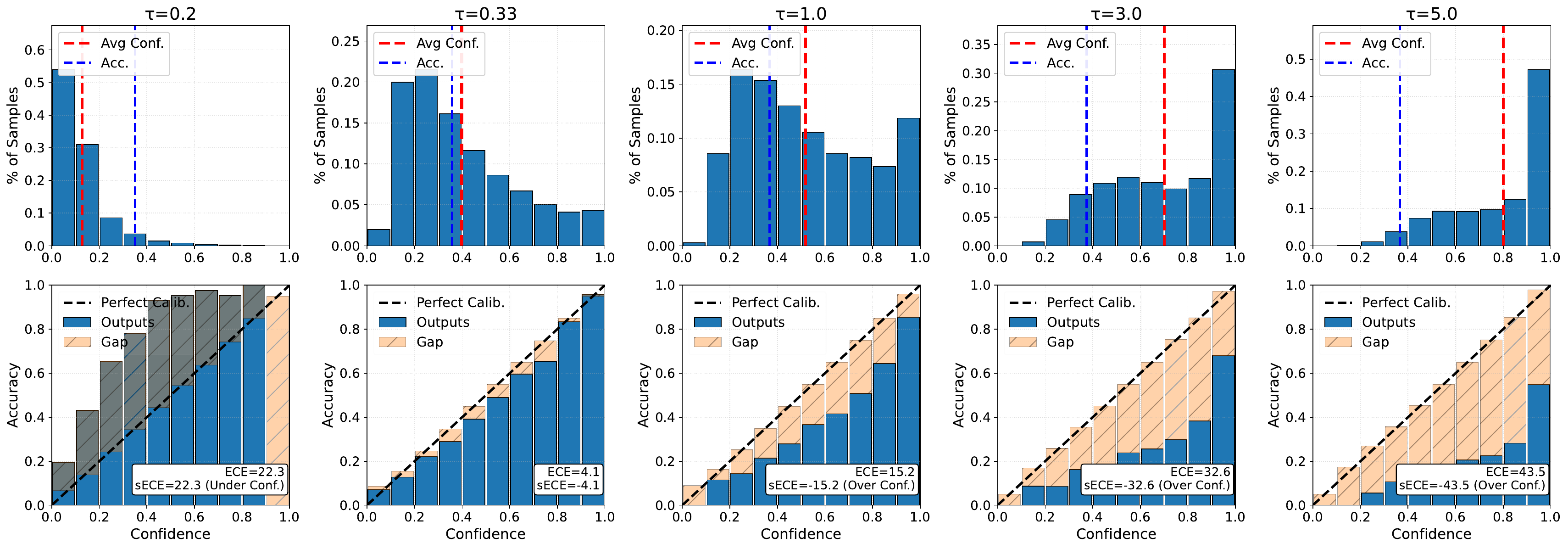}
        \caption{CIFAR100--ResNet18.}
        \label{fig:reliability-diagram-resnet}
    \end{subfigure}

    \begin{subfigure}[t]{\linewidth}
        \centering
        \includegraphics[width=0.85\linewidth]{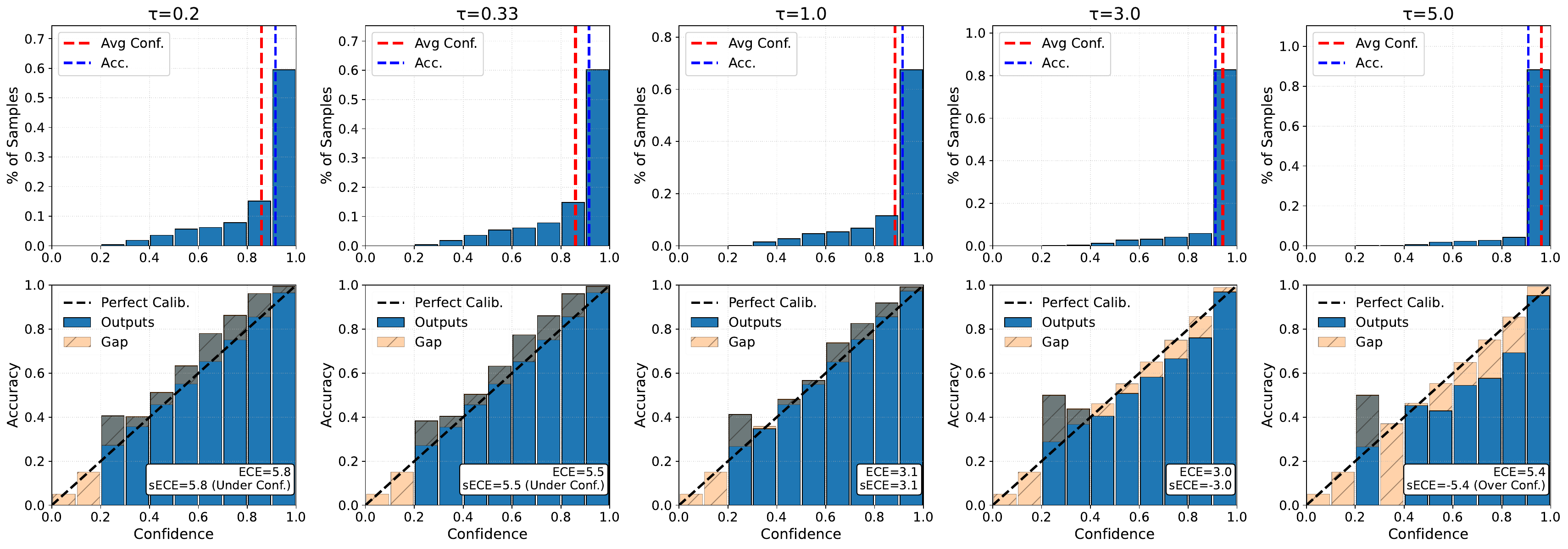}
        \caption{MNIST--MLP.}
        \label{fig:reliability-diagram-mlp}
    \end{subfigure}
    \caption{Additional confidence histograms and reliability diagrams under different training temperatures $\tau$.
    Across both settings, varying $\tau$ consistently shifts the confidence distribution and reliability curve while leaving accuracy largely stable.}
    \label{fig:reliability-diagram-additional}
\end{figure*}

\noindent{}\textbf{$\bullet$ Post-hoc calibration methods.}
For post-hoc calibration (Fig.~\ref{fig:posthoc-calibration}), we evaluate temperature scaling~\cite{guo2017calibration}, Platt scaling~\cite{platt1999probabilistic}, isotonic regression~\cite{zadrozny2002transforming}, and histogram binning~\cite{zadrozny2001obtaining}.
All methods fit a mapping from an uncalibrated model score to a calibrated confidence using a held-out calibration set.
\textbf{Temperature scaling}~\cite{guo2017calibration} fits a single scalar $T>0$ that rescales logits at inference, $\mathbf{p}=\mathrm{softmax}(\mathbf{z}/T)$, by minimizing negative log-likelihood on the calibration set. It preserves the predicted class (logit ordering) and only adjusts probability sharpness.
\textbf{Platt scaling}~\cite{platt1999probabilistic} fits a parametric sigmoid mapping (typically on a scalar score such as a logit margin), $p(y{=}1\mid s)=\sigma(as+b)$, via maximum likelihood on the calibration set. It is a two-parameter, monotone calibrator.
\textbf{Isotonic regression}~\cite{zadrozny2002transforming} fits a non-parametric \emph{monotone} piecewise-constant function that maps scores to probabilities by minimizing squared error subject to monotonicity. It is flexible but can overfit when the calibration set is small.
\textbf{Histogram binning}~\cite{zadrozny2001obtaining} partitions scores into bins and replaces each score by the empirical accuracy within its bin. This provides a simple non-parametric estimate of $P(\text{correct}\mid s)$, with a bias--variance trade-off controlled by the number of bins and calibration-set size.

\section*{Additional Experimental Results}
\noindent\textbf{Extended threshold-based triage analysis.}
Figure~\ref{fig:healthcare-threshold-extend} extends the mobile healthcare case study by plotting threshold-sweep behavior in percentage form. Using the same verification policy, we report how calibration distortion changes the rate of unnecessary verifications and missed verifications across confidence thresholds $\gamma$, illustrating the operational impact of under-/over-confidence beyond a single threshold.

\noindent\textbf{Additional Results for Robotics and Autonomous Driving (selective perception and safety gating).}
Figure~\ref{fig:detr-appendix} extends the DETR case study by reporting additional qualitative examples and summary statistics under different training temperatures. The results highlight how temperature-induced miscalibration shifts the confidence of the ``no-object'' class and, under a fixed score threshold, can lead to systematically different detection behaviors.

\noindent\textbf{Additional reliability diagrams.}
Figure~\ref{fig:reliability-diagram-resnet} reports the same confidence histograms and reliability diagrams as in Fig.~\ref{fig:reliability-diagram}, but for the CIFAR100--ResNet18 configuration (Seed~0). This serves as an additional qualitative check that the temperature-controlled calibration drift (under-/over-confidence as $\tau$ varies) persists in a harder, more miscalibrated regime.

Figure~\ref{fig:reliability-diagram-mlp} provides the corresponding visualization for MNIST--MLP (Seed~0). Although the task is simpler, the same pattern holds: changing the training temperature shifts the confidence distribution and the reliability curve in a consistent direction while leaving accuracy largely stable.

\end{document}